\documentclass[journal]{IEEEtran}
\usepackage{verbatim} %---to enable block commenting

\usepackage{blindtext, graphicx}

\usepackage[dvipsnames]{xcolor}

\usepackage{subfig}

\usepackage{wrapfig}
\usepackage{amsmath}
\usepackage{amsfonts}

\usepackage{graphicx}

\usepackage[english]{babel}
\begin{document}
\setlength{\abovedisplayskip}{5pt}
\setlength{\belowdisplayskip}{5pt}
%\begin{comment}
% Use-Cases for Connected and Autonomous Vehicles: V2X Communication Architecture, Requirements and Design Considerations

\title{Use of Parallel Explanatory Models to Enhance Transparency of Neural Network Configurations for Cell Degradation Detection}


\author{\IEEEauthorblockN{David Mulvey\IEEEauthorrefmark{1},
Chuan Heng Foh\IEEEauthorrefmark{1},
Muhammad Ali Imran\IEEEauthorrefmark{2}, and
Rahim Tafazolli\IEEEauthorrefmark{1}}\\
\IEEEauthorblockA{\IEEEauthorrefmark{1}5G/6G Innovation Center, Institute for Communications Systems, University of Surrey, Guildford, UK\\
Email: \{d.mulvey, c.foh, r.tafazolli\}@surrey.ac.uk}\\
\IEEEauthorblockA{\IEEEauthorrefmark{2}School of Engineering, University of Glasgow, Glasgow, UK\\
Email: muhammad.imran@glasgow.ac.uk}
}

%set up IEEE copyright notice
\IEEEoverridecommandlockouts
%\IEEEpubid{\makebox[\columnwidth]{978-1-5386-5541-2/18/\$31.00~\copyright2024 IEEE 
\IEEEpubid{\makebox[\columnwidth]{10.1109/TNNLS.2024.3373101~\copyright2024 IEEE
\hfill}
\hspace{\columnsep}\makebox[\columnwidth]{ }}

\maketitle
% required for copyright notice
\IEEEpubidadjcol

\begin{abstract} 
In a previous paper, we have shown that a recurrent neural network (RNN) can be used to detect cellular network radio signal degradations accurately. We unexpectedly found, though, that accuracy gains diminished as we added layers to the RNN. To investigate this, in this paper, we build a parallel model to illuminate and understand the internal operation of neural networks, such as the RNN, which store their internal state in order to process sequential inputs. This model is widely applicable in that it can be used with any input domain where the inputs can be represented by a Gaussian mixture. By looking at the RNN processing from a probability density function perspective, we are able to show how each layer of the RNN transforms the input distributions to increase detection accuracy. At the same time we also discover a side effect acting to limit the improvement in accuracy. To demonstrate the fidelity of the model we validate it against each stage of RNN processing as well as the output predictions. As a result, we have been able to explain the reasons for the RNN performance limits with useful insights for future designs for RNNs and similar types of neural network.
\end{abstract}


\IEEEpeerreviewmaketitle

\section{Introduction}

%\IEEEPARstart{T}{he}  

In the latest generation of cellular networks, 5G, the emergence of sophisticated new techniques such as large scale MIMO and multicarrier operation has resulted in rapid growth in the total number of radio access network (RAN) configuration parameters. This carries with it a considerable risk in terms of potential misconfiguration and is likely to significantly add to the workload for network management teams. Fortunately the recent emergence of powerful machine learning techniques has the potential to counter this by alerting operators to issues which might not otherwise be apparent and providing assistance to resolve them in a timely manner.

In our earlier work \cite{Mulvey2018}, we showed that it is possible to apply a recurrent neural network (RNN) to address an issue of particular concern to mobile network operators, namely how to detect cell performance degradations which are not being reported to the network control centre but are impairing the quality of service perceived by the users. Our solution achieved high accuracy and can be efficiently deployed under operational network conditions. We discovered, though, that adding layers to deepen the RNN resulted in diminishing gains in overall accuracy, and we were keen to understand why this was happening.

A limitation of the deep RNN approach, however, shared with other neural networks (NNs), is that the internal workings of the RNN are not transparent and it is difficult for users and even designers to understand how it processes the input data to arrive at a fault prediction. 

As we highlighted in \cite{Mulvey2019}, this problem is being addressed by the US Defense Advanced Research Projects Agency (DARPA) Explainable Artificial Intelligence (XAI) program \cite{DARPAEAIP2020}, which is
designed to enhance the transparency of deep NNs, especially those deployed in operationally critical applications. 

In this work, we are seeking to establish a framework in which to study a broad class of NNs, of which the RNN is one instance, which store their internal state in order to process sequential inputs. Applying the DARPA XAI principles, we have built a model to run in parallel with the RNN to illuminate the RNN's inner workings and enabling us to analyse its internal operation in some depth. An NN can be seen as a function approximator where the training process attempts to fit an arbitrary function to a set of given inputs. If the NN is used as, say a binary classifier, a threshold is defined at the output stage to separate the outputs into two predefined classes. Therefore, a well-trained NN for binary classification is equivalent to a function that manipulates and transforms the inputs to some outputs which can be separated by the threshold. Thus modelling a well-trained NN provides us insightful knowledge on how the NN attempts to manipulate the inputs for separation and whether any performance limit is introduced when doing so.
Although developed in the context of a cellular radio network, our work has the potential for wider applicability to any situation where the inputs to an RNN can be expressed as a Gaussian Mixture Model (GMM). In his analysis of this topic, Murphy \cite{Murphy2012} 11.2.1 states that: \emph{``Given a sufficiently large number of mixture components, a GMM can be used to approximate any density defined on $\mathbb{R}^D$''}. We have been able to obtain good results with a relatively small number of mixture components as we describe below.

Our approach is based on viewing the operation of the RNN from the probability distribution perspective. In other words,  the RNN transforms the probability density function (pdf) of the data at each stage of processing, with the positive outcome of increasing overall detection accuracy. We discovered, however, that this benefit comes with a negative effect namely the creation of side distributions, which act to limit the gain in overall accuracy. We note that the RNN internal processing includes non linear elements which we wish to linearise in order to keep the mathematics tractable. 

By developing linear approximations to these elements we were able to formulate a linearised model which is capable of closely tracking the RNN processing at each stage. We were then able to use this to analyse the operation of the RNN and explain why the performance limitations occur.

We have structured the current paper in the following way. Firstly, we review recent related work and explain how our approach to increase RNN transparency fits into the DARPA framework. In Section \ref{sec:modelling}  we explain the principles underlying the parallel model. In Section \ref{sec:system_design} we outline our system design and describe how we implemented it. In Section \ref{sec:validation}, we explain how we validated our model and discuss the results we obtained by applying it to a variety of RNN configurations. In Section \ref{sec:furtherwork} we make a number of suggestions for further work. Finally in Section \ref{sec:conclusion} we summarise our conclusions and draw out the implications for future RNN designs.

\section{Related Work}

There have been a number of recent papers reporting the application of NNs to the detection of faults in cellular networks. Zhang \emph{et al.}, for instance \cite{Zhang2020}, showed how generative adversarial neural networks can be used to synthesise realistic fault data in cases where limited volumes of such data can be captured from the network. In our earlier study \cite{Mulvey2018}, we used  an RNN to detect radio signal faults (transmission power reductions in dB)  which impact the quality of service offered to the users, but with current technology are not visible at the network operations centre. 

The simplest form of RNN \cite{Goodfellow2016}, which forms the baseline for our investigations, consists of a single layer operating on the current inputs and a feedback vector based on the previous internal state. We have extended this to cover two and three concatenated layers, each of the same basic type.

We also used a higher order extension of this architecture described by Zhang and Woodland \cite{zhang2018a} to experiment with a single layer with internal processing based on two and four previous states (referred to as second and fourth order RNNs respectively). 

\begin{figure}[!tb]
\centering
\includegraphics[width=8cm]{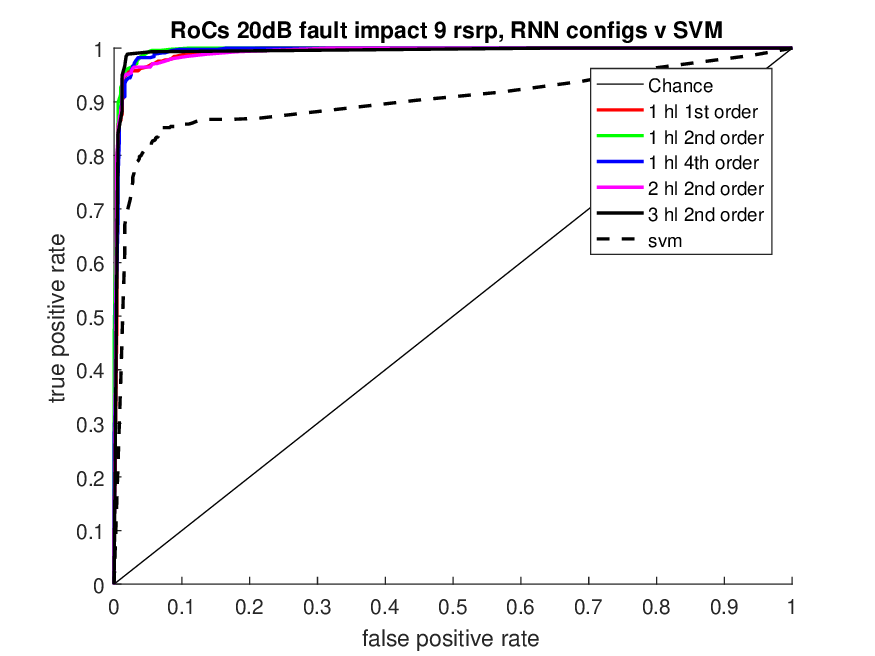}

\caption{RNN and SVM Results for fault impact of 20dB.}
\label{fig_rnr1}
\end{figure}

With this approach we found that we could significantly improve on the best results attainable from the use of earlier techniques such the Support Vector Machine (SVM) \cite{ZohaSaeedImranEtAl2015} (see Fig. \ref {fig_rnr1}), reproduced from \cite{Mulvey2018}. This shows the detection performance of each RNN configuration plus the SVM for a fault representing a 20dB loss in transition power. For the RNN configurations, the legend gives the number of RNN hidden layers followed by the RNN order as described above.

During this work we discovered, however, as shown in Fig. \ref {fig_rnr1}, that if we began with a single hidden layer and added more layers we encountered diminishing improvements in accuracy. We should note here that we are aware of the risks of overfitting to the training data, and hence reducing test accuracy, by adding too many layers or too many parameters per layer. In our case we were careful to keep the RNN configuration per layer to the minimum required to achieve the necessary result. 

A similar effect was observed when we increased the order of the internal processing to add earlier internal states to the feedback loop. This led us to study the internal operation of the RNN to understand why this was happening and as a result develop an explanation of how the RNN works. We published this work in our recent conference paper \cite{Mulvey2020}. 

As a result of the DARPA XAI programme \cite{Gunning2016}, \cite{Gunning}, there has been increasing interest in finding ways to increase the transparency of deep NNs especially in situations where decisions made by a deep NN may have significant operational or social consequences. 

The DARPA XAI launch document \cite{Gunning2016} proposes the following broad strategies to achieve deep NN transparency:
\begin{enumerate}
\item 
Develop modified or hybrid deep learning techniques that learn more explainable features, explainable representations, or explanation generation facilities
\item 
Develop alternative machine learning techniques that learn more structured, interpretable or causal models
\item 
Develop techniques that would experiment with any given machine learning model - as a black box - to infer an approximate, explainable model
\end{enumerate}

The "counterfactual" approach \cite{Karimi2020}, which treats the deep NN as a black box and attempts to discover what changes would need to be made to the inputs to achieve a desired output, is an example of the third strategy. A recent paper by Yang \emph{et al.} \cite{Yang2020} describes an implementation approach in line with the second strategy which accepts architectural constraints on the neural network in return for greater explainability. Our approach, by contrast, is a hybrid one and so under this scheme can be classified as an example of the first strategy. 
 
Since the DARPA strategies were published, however, a number of useful surveys \cite{Adadi2018a}, \cite{Tjoa2020}, \cite{Heuillet2020} and \cite{Arrieta2020} have charted recent progress in defining the issues in implementing XAI and classifying the approaches which have been taken to address these.

In particular, Arrieta \emph{et al} \cite{Arrieta2020} provide a useful taxonomy of recent XAI approaches. They distinguish between post hoc explanations, where the technique seeks to explain the black box output given the inputs, and approaches to enhance black box transparency. 

The authors subdivide post hoc explanations into text explanations, visual explanations, local explanations, explanations by example, explanations by simplification and feature relevance explanations. The integrated gradients method \cite{Sundararajan2017}, \cite{Peng2021} and \cite{Goh2021}, for example, would be an instance of feature relevance within the general category of post hoc explanations. 

Transparency approaches, on the other hand, are subdivided by the authors into simulatability (designing the model to be understood by humans, e.g. by keeping it simple), decomposability (designing the model so that the inputs, processing and outputs can be understood separately) and algorithmic transparency (designing the model so that the user can understand the process followed by the model to produce any given output from its input data). As an example of algorithmic transparency, Arrieta \emph{et al} state that: \emph{``A linear model is deemed transparent because its error surface can be understood and reasoned about''}.

So it will be seen that by contrast with post hoc explanations such as integrated gradients, we are seeking to increase the transparency of the black box, in our case by building a parallel linear model which is transparent to human review. In our work, rather than seeking to explain the outputs given the inputs, our objective is rather to understand the mechanics of the system and its performance tradeoffs. Hence we are following the algorithmic transparency subcategory defined by Arrieta et al, by studying the internal workings of the RNN in depth. 

To do this, we utilise a tractable pdf to approximate the input features and build a parallel linearised model whose intermediate states can be checked against those of the RNN. This allows us to study in detail how the RNN reshapes the input distributions prior to the final detection decision. This gives us a number of key insights into the internal workings of the RNN, leading to an improved understanding of how to achieve an appropriate balance between RNN output accuracy and resource utilisation.  

An important limitation of some post hoc explanations such as the integrated gradients approach is that they assume that the network output is a function of the current inputs only \cite{Sundararajan2017}. For networks such as the RNN, however, where the NN includes memory elements which are influenced by previous inputs, it is possible that the transfer function for a given current input will vary depending on the previous inputs. In this situation this type of approach may give invalid or unstable results. By contrast our approach, which takes into account the temporal processing of the NN, can correctly process input sequences of arbitrary length. 

\begin{figure*}[!t]
\centering
\includegraphics[width=16cm]{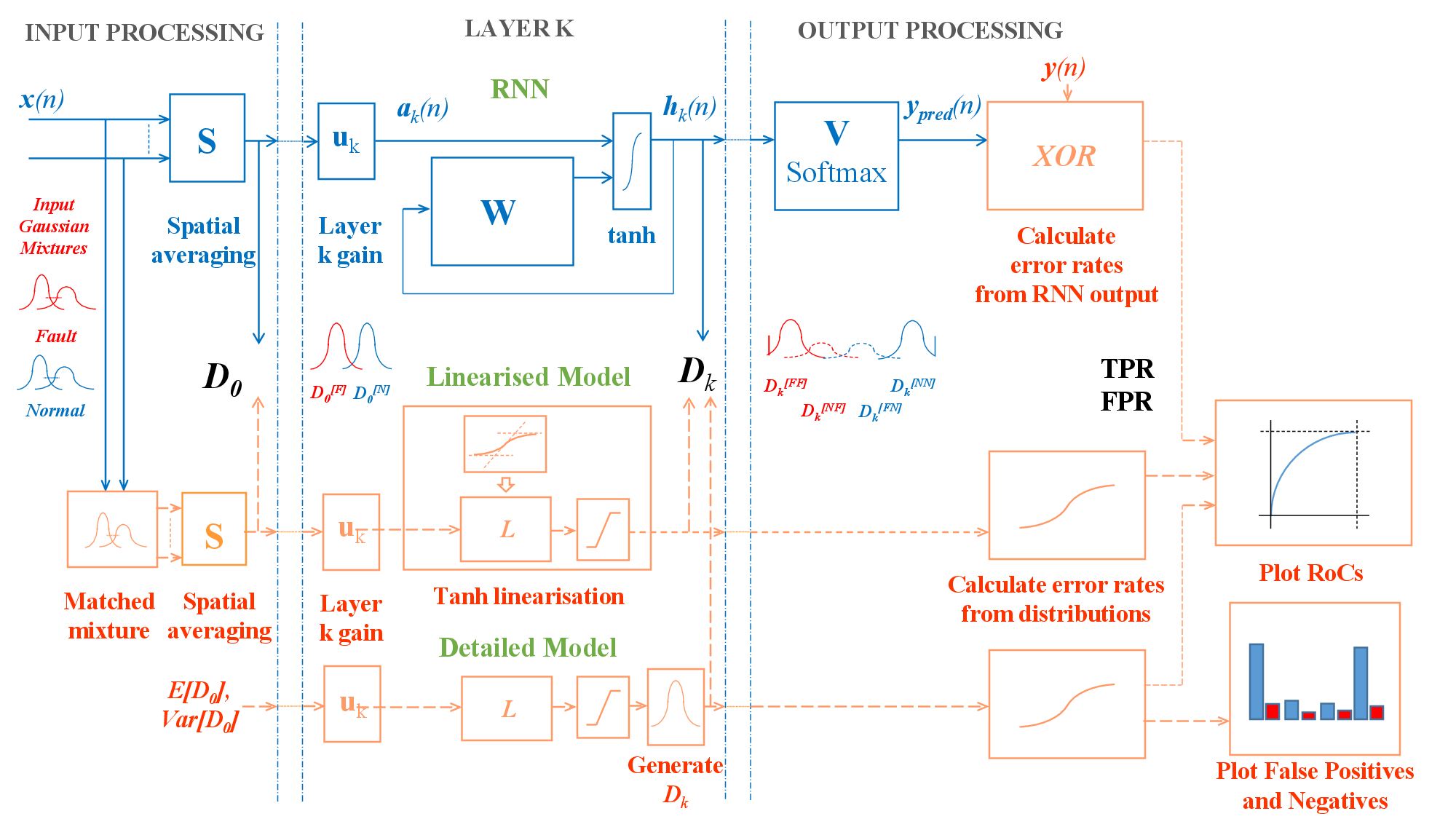}

\caption{System Diagrams of RNN and Models Showing Parallel Process Flow.}
\label{fig_qlm}
\end{figure*}

\section{Modelling Approach} \label{sec:modelling}

Our earlier work in \cite{Mulvey2018} has demonstrated the application of a RNN for cell degradation detection, with RSRP measures as inputs. In this paper, we focus on developing an improved understanding of RNN performance behaviour. As mentioned earlier, we view training the RNN as an input remapping process, enabling the RNN to transform inputs in order to separate different categories of input data for more accurate classification. Hence modelling a well-trained RNN allows us to understand its performance behaviour. Such a study needs to model both the input data and also the RNN internal operation. For the this application, our inputs are RSRP measures which can be characterised by a mixture of Gaussian distributions. The model we have developed describes how these inputs undergo transformation within the RNN to achieve greater separation between categories at the input to the classifier stage.

Figure \ref{fig_qlm} depicts our linearised model of a multi-channel RNN (of arbitrary hidden layer width), with the input features represented as samples from a Gaussian mixture. Without loss of generality, we consider a particular hidden layer channel and compare each stage of internal processing in the model with the equivalent in the RNN. We validate the model's outputs against those of the RNN by plotting Receiver Operating Curves (RoCs) and also by comparing the output distributions (see Figures \ref{fig:1storder123layers_D4x_RoCs_RNN} and \ref{fig:1layer124orders_D4x_RoCs_RNN}). We supplement the linear model with a more detailed model based on the same principles, allowing us to study in depth each component of each intermediate distribution and quantify its impact on the overall detection accuracy. 
\subsection{Input Data}

The input data are taken from a system simulation, which is described in more detail later in this paper. The simulation represents a dense deployment consisting of a ring of cells around a central cell, plus a group of users who move around the simulated area according to a constrained random walk. We apply faults to the central cell only. 

Each user provides a periodic report on the Reference Signal Received Power (RSRP) received from each base station. The input features consist of a set of these reports, taken at the same time period, giving the RSRP values which the user is currently measuring for the central cell. These reports are structured according to sequences of a configurable constant length, during which we apply a fault at a random point and then allow it to persist until the end of the sequence. We also provide a label stream indicating whether a fault is currently present or not, which is used to train the RNN.

We selected a total of 9 input features representing a typical number of users active at any given time based on the chosen deployment scenario; the RNN configuration is able to handle as many additional users as are likely to be required for this scenario.

The RSRP value is expressed in dB and includes the effects of antenna gain, pathloss, log-normal fading and fast fading.

\subsection{Process Building Blocks}

We analyse the RNN configurations we utilised in \cite{Mulvey2018}, concentrating initially on the single layer, first order case (see Figure \ref{fig_qlm}). We then apply these principles to cover the multi-layer and higher order cases. We can divide the RNN operation into three principal stages. Stage 1 covers the weighted aggregation of multiple RSRP reports from user equipment at different locations, which we refer to as spatial averaging. Stage 2 is based on a representative hidden layer within the RNN, and includes the recurrent element of the processing, based on feedback of the internal state, which we designate as temporal processing. Stage 3 comprises the output processing including the fault detector.
 
We assume that each hidden layer has more than one channel so we represent the state of the $kth$ hidden layer by the vector $\textbf{h}{_k}(n)$ as shown on Figure \ref{fig_qlm}. 

In this section we explain how the RNN input features are represented, followed by how spatial averaging is treated within our models, and finally how we approximate the temporal processing part of the RNN. Output processing is straightforward in that it can be represented by a simple threshold so is not discussed further.

\subsubsection{Input Distributions}
We apply the Gaussian Mixture approach \cite{Murphy2012} and model the input RSRP values by a one-dimensional Gaussian mixture where the $i$-th example has the probability distribution:
\begin{equation} \label{eq:gmm1}
p(x_{i}) =  \sum\limits_{k=1}^K w_{k} \mathcal{N}(x_{i}|\mu_{k},\sigma_{k})
\end{equation}

where
\begin{equation} \label{eq:gmm2}
 \mathcal{N}(x_{i}|\mu_{k},\sigma_{k}) = \frac{1}{\sigma_{k}\sqrt{2\pi}}exp\left(-\frac{(x_{i}-\mu_{k})^2}{2\sigma_{k}^2}\right)
\end{equation}

and with the constraint that:

\begin{equation} \label{eq:gmm3}
\sum\limits_{k=1}^K w_{k} = 1.
\end{equation}

By exploiting the property of Gaussian distributions of remaining closed under the addition operation \cite{Weissteinb},\cite{Weisstein},\cite{Weissteina}, we can use simple linear operations to calculate the mean and standard deviation of each subcomponent distribution at each stage of RNN processing, except where the processing contains a non-linear element, for which we need a linear approximation as we describe below.

\subsubsection{Spatial Averaging}
We first of all establish the distribution transfer function relating the output and input distributions, and we then discuss the effect of this stage on reshaping the input distributions.

We notice that the input processing stage of the RNN in our scenario is performing weighted spatial averaging where the input feature set consists of RSRP measures reported by users from a range of different locations across the coverage area of the cell of interest. Each feature is periodically updated with an RSRP measure from one allocated user at a varying location, then the RSRP measures from all the input features are consolidated using weighted averaging. 

At any given moment the network is either operating normally or a fault condition is present. The spatial averaging stage applies matrix $\textbf{S}$ to map the input features to the first hidden layer. Consider the effect on the $c$-th hidden layer channel of spatial averaging across all the RSRP values at a given instant when the network is operating normally (the faulty case follows similar logic).

Let $x_i(n)$ be the value of the RSRP signal reported by the $i$-th user, such that $i=1,2,...,m$, at instant $n$, and denote the sample set at this instant as $X(n)$. Each sample $x_i(n)$ is taken from one or other element of the Gaussian mixture at that instant. If we assume that the selection at any one instant is random, we can say that the probability of each sample belonging to the $k$-th element of the mixture is $w_k$. Define a vector 
\begin{equation} \label{eq:mnl1}
\mathbf{q} = (q_1,...q_K)
\end{equation} where $q_k$ is the number of times the $k$-th component of the Gaussian mixture appears in the sample set. The probability of any given composition of the complete sample set can be modelled by a multinomial distribution of category size $K$ and number of trials $m$ 
with a probability mass function, {$f_X(m,K)$}, given by \cite{Murphy2012}:

\begin{equation} \label{eq:mnl2}
f_X(m,K) = \left(\frac{m!}{q_1!q_2!...q_K!}\right)\prod\limits_{k=1}^Kw{_k^{q_k}}
\end{equation}

For any given composition of the sample set, to average the components we add them together weighted by the $c$-th row of $\mathbf{S}$. Since each component is normally distributed, the addition property for normal distributions \cite{Weisstein} states that this will result in a single distribution. The above theory predicts that this will mean we will see a family of Gaussian distributions, one for each of the compositions of the set represented by the multinomial distribution. 

What we see in practice, however, is that we obtain a single bell shaped distribution, either from the RNN itself or when we model the Gaussian mixture and average it in the linear model. We can readily fit a single Gaussian distribution to this, and we use this for the detailed model. We designate this distribution as   $D_{0,c}$, with subcases $D_{0,c}^{[N]}$ for normal operation and $D_{0,c}^{[F]}$ for the case when a fault is present.

\begin{figure}[!htb]
\centering
\includegraphics[width=9cm]{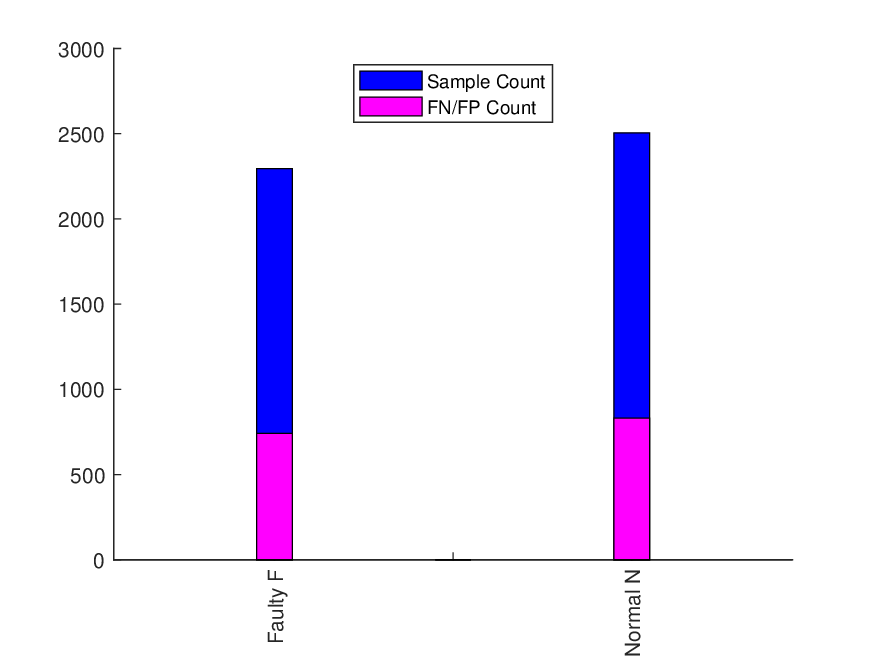}
\caption{Illustration of RNN Inputs (15dB impact).}
\label{Fig_N12bar}
\end{figure}

\begin{figure}[!htb]
\centering
\includegraphics[width=9cm]{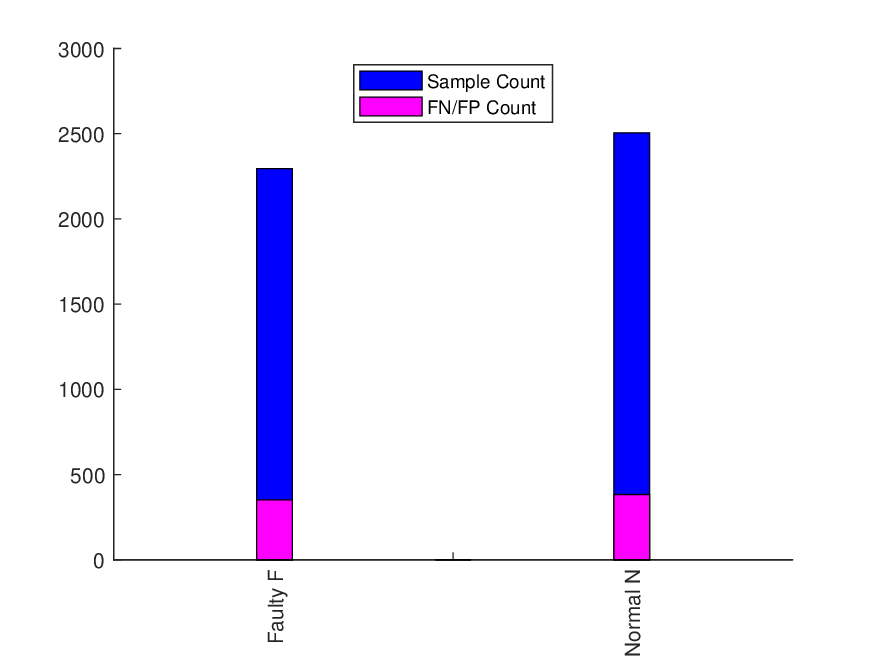}
\caption{Illustration of $D_0$ per channel (15dB impact).}
\label{Fig_N3bar}
\end{figure}

In other words, spatial averaging reshapes the input distributions by transforming the input Gaussian mixture into a single distribution which can be closely approximated by a Gaussian distribution $D_{0,c}^{[N]}$ (resp. $D_{0,c}^{[F]}$) per hidden layer channel. The most important effect of this is to decrease the overall distribution variance. As a result the area in common between the normal and fault cases is decreased and so the false positive and negative counts are reduced. The result of this is shown in Figures \ref{Fig_N12bar}, representing the input to spatial averaging and \ref{Fig_N3bar}, representing the output of this stage for a representative channel. In Figure \ref{Fig_N3bar} we can see that for $D_0^{[N]}$ and $D_0^{[N]}$ the combined count of false positives and false negatives has fallen compared with the same count in Figure \ref{Fig_N12bar}, and hence the proportion of samples at the output which would be incorrectly classified is smaller than at the input.    

\subsubsection{Temporal Processing}

As with spatial averaging, we derive the distribution transfer function and then discuss the reshaping effect of this stage. 

Time or sequence related processing, however, is more complex than the previous stage. As this part of the RNN contains a non-linear function (NLF), to accommodate the modelling while avoiding unnecessary complexity, we approximate the process using a linear model.  To achieve a linear model we need to (i) linearise the NLF which in our case is the tanh$(\cdot)$ function and (ii) re-express the internal state feedback matrix as a transformation operating exclusively on the temporal stage inputs. We linearise the NLF with a series of line segments defined by a gain $g(\cdot)$ and an offset $r(\cdot)$. For any given sample the line segment in use will depend on the value of the NLF input. The outcome of the $h(\cdot)$ feedback transformation, however, is that the linearised NLF output depends on a sequence of NLF inputs and hence a sequence of different $(g(\cdot)$,$r(\cdot)$) pairs, which we call the line segment sequence (LSS). This sequence will be different for each hidden layer channel. We denote the set of these sequences as the LSS set $\mathbf{L}$. 

In Figure \ref{fig_qlm}, we see that due to the feedback in each RNN layer, $h_k(n)$ which is the output from layer $k$ depends on $a_k(n)$, and $h_k(n-1)$. With further analysis (given in Appendix B), we show that the dependencies on $a_k(n)$ and $h_k(n-1)$ can be transformed into one on the current and previous states of $a_k(\cdot)$ only, using the LSS parameters $g(\cdot)$ and $r(\cdot)$. In \eqref{eq:nlflin_09c}, \eqref{eq:nlflin_15} and \eqref{eq:nlflin_16}, we show that for a first order system, if we approximate this result by considering $a_k(n)$, $a_k(n-1)$ and $a_k(n-2)$ only, $\mathbf{h}(n)$ is given by:

\begin{equation} \label{eq:nlflin_09b}
\textbf{h}(n) 
  = \boldsymbol{\alpha}{_\textbf{0}}{^{\textbf{[L]}}} \mathbf{a}(n)
   +\boldsymbol{\alpha}{_\textbf{1}}{^{\textbf{[L]}}} \mathbf{a}(n-1)
   +\boldsymbol{\alpha}{_\textbf{2}}{^{\textbf{[L]}}} \mathbf{a}(n-2)
   +\boldsymbol{\beta}{^{\textbf{[L]}}}
\end{equation} 
with

\begin{equation} \label{eq:tp_nlflin_01}
\begin{array}{l}
\boldsymbol{\alpha}{_\textbf{0}}{^{\textbf{[L]}}} = \textbf{g}{_\textbf{0}} \\
\boldsymbol{\alpha}{_\textbf{1}}{^{\textbf{[L]}}} = \textbf{g}{_\textbf{0}} \odot  \mathbf{W_1}\textbf{g}{_\textbf{1}} \\
\boldsymbol{\alpha}{_\textbf{2}}{^{\textbf{[L]}}} = \textbf{g}{_\textbf{0}} \odot \textbf{g}{_\textbf{1}} \odot  \mathbf{W_1} \mathbf{W_1} \textbf{g}{_\textbf{2}} \\
\end{array}
\end{equation}
and
\begin{equation} \label{eq:tp_nlflin_02}
\boldsymbol{\beta}{^{\textbf{[L]}}} = \textbf{r}{_\textbf{0}} + \textbf{g}{_\textbf{0}} \odot \mathbf{W_1} \textbf{r}{_\textbf{1}} + [\textbf{g}{_\textbf{0}} \odot \mathbf{W_1}\mathbf{W_1}\textbf{g}{_\textbf{1}}]\textbf{r}{_\textbf{2}}
\end{equation}
where the LSS set $\textbf{L}$ is given by: 
\begin{equation*}
\textbf{L} = \langle (g_0,r_0),(g_1,r_1),(g_2,r_2) \rangle. 
\end{equation*}
and $\boldsymbol{\alpha}{_\textbf{0}}{^{\textbf{[L]}}}$, $\boldsymbol{\alpha}{_\textbf{1}}{^{\textbf{[L]}}}$, $\boldsymbol{\alpha}{_\textbf{2}}{^{\textbf{[L]}}}$ and $\boldsymbol{\beta}{^{\textbf{[L]}}}$ are constant for this LSS. We note that in this case, where we are considering a first order RNN, the length of the LSS is three pairs of $g_n$ and $r_n$ values. For the second order and fourth order cases the length becomes five and nine pairs respectively, as we discuss later on in this paper.

If we consider the first layer, layer 1,  we note that for channel $c$ of the hidden layer we have:
\begin{equation} \label{eq:a1}
a_{1,c}(n)\ = u_{1,c} d_{0,c}(n)
\end{equation}
where $d_{0,c}(n)$ is a sample taken from $D_{0,c}$  and 
\begin{equation} \label{eq:n4}
d_{1,c}(n)\ = h_{1,c}(n)
\end{equation}
where $d_{1,c}(n)$ is a sample taken from $D_{1,c}$ which is the distribution of channel $c$ at the output of layer 1. For clarity we drop the subscript $c$ on the understanding that the following analysis applies on a per channel basis.

Hence combining \eqref{eq:tp_nlflin_01}, \eqref{eq:a1} and \eqref{eq:n4}, we arrive at the equation representing the end to end processing of the layer: 
\begin{equation} \label{eq:tanhlin_6}
\begin{array}{ll}
d_{1}(n) = & u_1\left[\alpha_0^{[L]}d_{0}(n) + \alpha_1^{[L]} d_{0}(n-1)\right. \\
& \left. \;\;\;\;\; + \alpha_2^{[L]} d_{0}(n-2)\right] + \beta^{[L]}.
\end{array}
\end{equation}

For the $k$-th layer such that $k\neq 1$ the above becomes:
\begin{equation} \label{eq:tanhlin_7}
\begin{array}{ll}
d_{k}(n) =  & u_k \left[\alpha_0^{[L]}d_{k-1}(n) + \alpha_1^{[L]} d_{k-1}(n-1)\right. \\
& \left. \;\;\;\;\; + \alpha_2^{[L]} d_{k-1}(n-2)\right] + \beta^{[L]}. 
\end{array}
\end{equation}

As we have seen, the RNN temporal processing operates on a sequence of inputs. We now show that the effect of the temporal stage operating on this sequence is to transform the spatial averaging output from two single distributions $D_0^{[N]}$ and $D_0^{[F]}$ to two main distributions and plus a set of unwanted distributions. We call these the main and side lobes respectively. 

Equation \eqref{eq:tanhlin_6} takes as its input a sequence of samples (in this case three) taken from $D_{0}$ at successive sampling instants, which we call the Fault Status Sequence (FSS). If we designate the samples indicating a network fault status of ``normal" by $N$ and those indicating a network fault status of ``faulty" by $F$,  there are eight possible FSSs: $NNN$, $NNF$, $NFN$, $NFF$, $FNN$, $FNF$, $FFN$ and $FFF$. More generally, for n samples there will be $2^n$ possible sequences. We denote the set of these by $FSS\_n$.

We now study each element of the specific set $FSS\_3$ and derive the corresponding layer 1 outputs. Consider first the $NNN$ case, where each of the three successive samples indicate fault status "normal", or in other words each sample is taken from the distribution $D_0^{[N]}$. For this case, we get:
\begin{equation} \label{eq:discn_fo3m}
E[D_{1}^{[NNN]}] = u_1[\alpha_0^{[L]} + \alpha_1^{[L]} + \alpha_2^{[L]}]E[D_0^{[N]}]] +\beta^{[L]}
\end{equation} 
and
\begin{equation} \label{eq:discn_fo3v}
Var[D_{1}^{[NNN]}] = u_1^{2}[(\alpha_0^{[L]})^2 + (\alpha_1^{[L]})^2 + (\alpha_2^{[L]})^2] Var[D_0^{[N]}] 
\end{equation}
For the $FFF$ case, we obtain a similar result, this time based on $D_0^{[F]}$.
We note that in both of these cases the mean is displaced by $\beta^{[L]}$, so that the overlap between the distributions is unaffected. We adjust each mean by subtracting  $\beta^{[L]}$ and then calculate the adjusted mean to SD ratio. The $u$ scaling factor does not impact this ratio, so for both $NNN$ and $FFF$ the adjusted mean to SD ratio increases by the multiplier $\frac{\alpha_0^{[L]} + \alpha_1^{[L]} + \alpha_2^{[L]}}{\sqrt{(\alpha_0^{[L]})^2 + (\alpha_1^{[L]})^2 + (\alpha_2^{[L]})^2}}$
, reducing the common area between $D_{1}^{[NNN]}$ and $D_{1}^{[FFF]}$ relative to that between  $D_0^{[F]}$ and $D_0^{[N]}$, resulting in a reduced error count.

In situations where both $N$ and $F$ are present in the FSS, however, we encounter a different effect. Considering, say, the $NNF$ case, we find that
\begin{equation} \label{eq:discn_fo9m}
\begin{aligned} 
E[D_{1}^{[NNF]}] = {} & u_1[\alpha_0^{[L]} E[D_0^{[F]}] 
+\alpha_1^{[L]} E[D_0^{[N]}] \\
& +\alpha_2^{[L]} E[D_0^{[N]}]] 
+ \beta^{[L]} 
\end{aligned} 
\end{equation}
\begin{equation} \label{eq:discn_fo9v}
\begin{aligned}
Var[D_{1}^{[NNF]}] = {} & u_1^{2}[(\alpha_0^{[L]})^2\ Var[D_0^{[F]}]+ (\alpha_1^{[L]})^2\ Var[D_0^{[N]}] \\
 & + (\alpha_2^{[L]})^2\
 Var[D_0^{[N]}]]
\end{aligned}
\end{equation}

The resulting distribution is still Gaussian, but now the mean is based on the weighted sum 
of $E[D_0^{[N]}]$  and $E[D_0^{[F]}]$. Given that $Var[D_0^{[F]}] = Var[D_0^{[N]}]$, the variance of $D_{1}^{[NNF]}$, however, is the same as the $NNN$ case. The other cases where the fault status indication varies during the fault status sequence lead to similar results. The effect is that the distributions for these cases have mean values which fall between those for $NNN$ and $FFF$.The relative peak magnitude of each of these in comparison  with $D_{1}^{[NNN]}$\ and $D_{1}^{[FFF]}$\ is determined by their relative frequency of occurrence. 
 
We refer to the distributions due to the $NNN$ and $FFF$ cases as the main lobes and the intermediate distributions as the side lobes. Each of these distributions arises from a single element of the input FSS. Table \ref{tbl:N4xxx_relfreqs} gives typical key parameters for these distributions for the sequence $FSS\_3$. The table shows that the main $NNN$ and $FFF$ cases have by far the highest sample counts, whereas some intermediate cases, for example $NFN$ and $FNF$,are relatively infrequent. Hence we can focus on the two main cases plus the most frequent intermediate cases, which we call the ``principal sidelobes", neglecting the infrequent intermediate cases. We can express this as a simple rule: we include those intermediate cases which have one transition from $N$ to $F$ or $F$ to $N$ and discard the others. This assumes that the fault occurs no more than once per FSS, and that the minimum fault duration exceeds or is the same as the length of the FSS.

The effect of the sidelobes, as we shall see, is the contribution of additional classification errors acting to reduce the improvements in main lobe accuracy which would otherwise be achieved by adding layers or moving to higher order RNN configurations.

\subsection{Multiple Layers}

\begin{figure}[!htb]
\centering
\includegraphics[width=9cm]{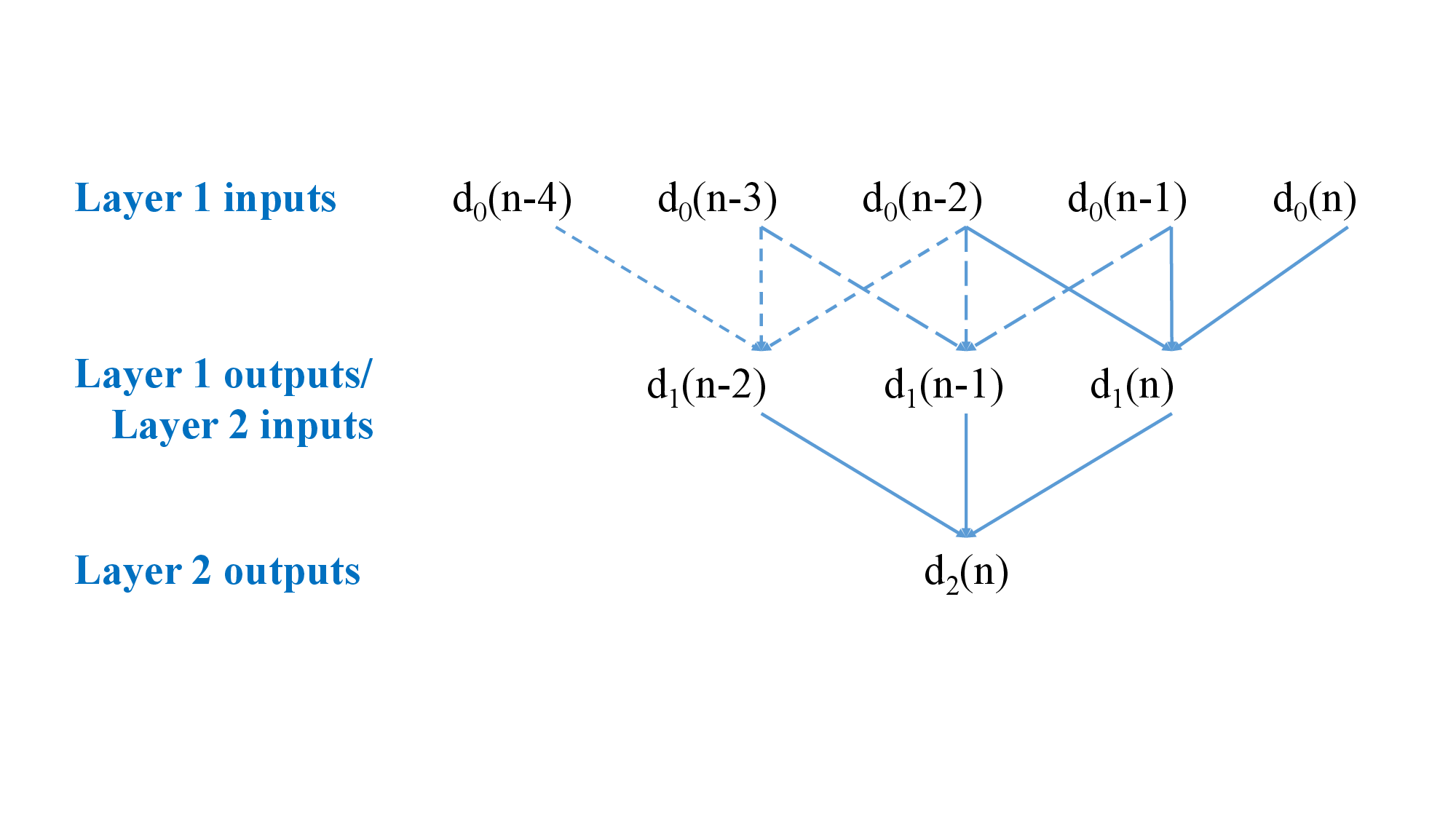}
\caption{Relationship between Layer 2 and Layer 1 processing}
\label{Fig_L1L2Deps}
\end{figure}

Consider Layer 2 processing in the case where both Layer 1 and Layer 2 are first order. As we have discussed, each Layer 1 output depends on three previous layer 1 inputs. The same applies to Layer 2. It follows that the current Layer 2 output depends on the current Layer 1 output and the previous two outputs, each of which have the same relation to the Layer 1 inputs as shown in Fig. \ref{Fig_L1L2Deps}. Hence the current Layer 2 output depends on the current and four previous Layer 1 inputs.

This means that for Layer 2 we now have an FSS consisting of five consecutive inputs. We denote the set of possible combinations of these as FSS\_5.

As an example we look at the situation in which the current Layer 1 output is based on the Layer 1 input FSS $NNF$. 

When the current layer 1 output was calculated, the current layer 1 input would have been $N$, the previous one $N$ and the one prior to that $F$. When the two earlier layer 1 outputs were calculated, the layer 1 FSSs would have been $x_3NN$ and $x_4x_3N$, in which $x_3$ and $x_4$ represent the layer 1 inputs which are respectively 3 and 4 samples prior to the current one. This leads to four alternative length 5 FSSs: $NNNNF$, $NFNNF$, $FNNNF$ and $FFNNF$. Counting the number of occurrences for the three with multiple transitions, we find that these are infrequent, confirming that based on the above rule we can neglect them to leave just $NNNNF$. The same logic can be applied to the other cases where the Layer 1 FSS contains both $N$s and $F$s. 

In the situation where the Layer 1 FSS contains only one fault status, however, we find that this leads to multiple cases. So for example the $NNN$ Layer 1 input FSS could form part of any of the length 5 input sequences $NNNNN$, $NFNNN$, $FNNNN$ or $FFNNN$. Of these only the $NFNNN$ case can be neglected, resulting in a total of 3 cases. The same logic can be applied to the $FFF$ case.

This means that at the Layer 2 output there are two cases with only one fault status namely $NNNNN$ and $FFFFF$, which become the main lobes at this point. There are now, however, a total of eight cases containing both $N$s and $F$s. The result of this is that at the output of Layer 2 we still have two main distributions but we now obtain eight of the principal sidelobes rather than the four at the output of Layer 1.

For a third layer, using the same logic we can demonstrate that for first order configurations we need to take into account a Layer 1 FSS of length 7, for which we denote the set of possible sequences as $FSS\_7$. For this scenario we obtain a total of twelve principal sidelobes alongside the two main lobes. 

In general, the ``single transition rule" above means that for an FSS of length $l_{FFS}$ there will be $2(l_{FFS}-1)$ principal sidelobes. For multilayer scenarios, the $k$-th layer has input sequence length $2k+1$ and at the output of this layer we obtain $4k$ principal sidelobes. 

Each of these additional sidelobes acts as a source of further classification errors, so that the overall effect is to reduce the gain in accuracy which would otherwise be achieved by adding more layers.
  
\begin{table}[!b] 
%% increase table row spacing, adjust to taste
\renewcommand{\arraystretch}{1.3}
% if using array.sty, it might be a good idea to tweak the value of
% \extrarowheight as needed to properly center the text within the cells
\caption{Key Parameters for $D_1$ distributions arising from FSS\_3 (Sequence Length 20, Fault Impact 15dB) }
\label{tbl:N4xxx_relfreqs}
\centering
{\begin{tabular}{|c|c|c|c|c|}

\hline
Case&Mean & SD  &Rel. Freq. &No. of Samples\\
\hline
FFF&-1.64 &1.10&0.41  & 1996 \\
\hline
NFF&-1.04&1.10& 0.04  & 203 \\
\hline
FNF &-0.63&1.10& \textless0.01 & 8 \\
\hline
NNF &-0.04&1.10& 0.04 & 210 \\
\hline
FFN &0.07&1.10& 0.04 & 203 \\
\hline
NFN &0.66&1.10& \textless0.01 & 15 \\
\hline
FNN &1.07&1.10& 0.04 & 210 \\
\hline
NNN &1.66&1.10& 0.42 & 1955 \\
\hline
Total &-&-& 1.0 & 4800 \\
\hline
\end{tabular}}

\end{table}

\subsection{Higher Order RNNs}

To obtain a broad picture of the effects of increasing order on detection accuracy, we studied first, second and fourth order RNN configurations. A third order configuration would follow the same principles.

We have already covered the first order case in the multiple layers example. In the case of the second and fourth order configurations, we show in Appendix B that the number of coefficients in the linearisation sequence increases from 3 in the first order case to 5 in the second order case and 9 for the fourth order configuration.

Hence using the same terminology as for multilayer systems, for second order we have a set of input cases which we again denote as $FSS\_5$ and for fourth order we have a set of input cases which we denote as $FSS\_9$. As before, the purely normal or faulty cases result in main lobe distributions whereas those in which the fault statuses of the inputs are not the same lead to the generation of sidelobes.

As before, we can reduce the number of cases by applying the "single transition rule" given above, which allows us to neglect the less frequent sidelobes with multiple transitions. As before, there are eight principal sidelobes for $FSS\_5$ and applying the formula above we get sixteen principal sidelobes for $FSS\_9$.

As with the multilayer case, these additional sidelobes act as a source of further classification errors, reducing the gain in accuracy which would otherwise be achieved by moving to a higher order configuration.

\section{System Design and Implementation}\label{sec:system_design}

\subsection{Network Simulation}

\begin{table}[!t]
\small
%% increase table row spacing, adjust to taste
\renewcommand{\arraystretch}{1.3}
\caption{Configuration of Simulator}
\label{table_scd}
\centering
\begin{tabular}{|p{2cm}|p{6cm}|}
\hline
\textbf{Simulator}\newline \textbf{Element}  & \textbf{Configuration}\\
\hline
eNodeBs & Hexagonal grid\newline One eNodeB at centre of ring of six\newline Distance between eNodeBs 200m\newline Transmit power 33dBm \\
\hline
Radio \newline Propagation & Pathloss model 3D TR36.873 \cite{TR368732017}

Channel model 3D-UMi\newline  Width of street 20m, height of buildings 20m 
\newline Transmitter height 10m Receiver height 1.5m \\
\hline
Radio Link & Carrier 2000MHz, Tx B/w 10MHz \newline 4 Tx and 4 Rx antennas\newline Tx  Mode 4 (Closed Loop Spatial Multiplexing)   \\
\hline
UEs & 21 \\
\hline
User Walking Model & 
Random walk, maximum excursion 100m  \\

\hline
\end{tabular}
\end{table}

We have based our network simulation on the Technical University of Vienna (TUV) LTE system level simulator, implemented in MATLAB. Our network scenario is based on a macrocell-small cell split architecture with the macrocell and the small cells operating on different frequencies. We assume the network is fully planned and we model the small cells only. We use the  3GPP TR36.873 \cite{TR368732017} 3D propagation model. From this we select the UMi scenario, rather than one of the UMa scenarios also defined in the report, on the basis that this is appropriate to our urban dense deployment scenario with an inter-eNodeB spacing of 200m.

We model seven small cells, six of which are configured as a ring with the seventh cell in the middle. We apply faults to the middle cell only. We model pedestrian users, with each user positioned in one of the small cells at the start of the simulation. We generate the subsequent user trajectories using a controlled excursion random walk algorithm. The exact approach to user trajectory generation does not appear to be critical in that even a purely random selection of user location resulted in an RSRP distribution which could be fitted to a similar Gaussian mixture to the one used in our study.

We have extended the simulator code to calculate Reference Signal Received Power (RSRP) levels as per the definition in \cite{TS362142022} 5.1.1, which is based on the Cell Specific Reference (CRS) signals defined in \cite{TS362112022} 6.10.1. We use these to  produce simulated Minimise Drive Testing (MDT) reports  \cite{TS373202022} and  \cite{TS363312022} which are fed into the RNN fault detector. 

\subsection{RNN Fault Detector}

We have produced code to support a variety of RNN configurations, permitting feedback as far back as four previous values of the internal state, and also with the ability to handle multiple hidden layers. We measure prediction accuracy by working out the rates of true positive and negative (TP/TN) predictions and similarly those of the false positive and negative (FP/FN) predictions. 
 
The following RNN configurations were implemented:
\begin{enumerate}
\item 
one hidden layer, with feedback of one, two, or four previous internal states
\item 
two hidden layers, with feedback of one previous state at each layer 
\item 
three hidden layers, with feedback of one previous state at each layer
\end{enumerate}

For each of these configurations the input is structured as a set of examples, each of which consists of a time series of 20 samples. For each example the fault condition is applied at a random point within the time series, selected independently of any previous examples, and persists until the end of the series. The RNN training data set contains 144 examples whereas the validation and training data sets each comprise 48 examples. 

\subsection{Modelling Implementation} 

We have implemented two models. The first is the main linearised model, which is used to validate the linear mathematical approximations and provide insights into the internal processing of the RNN. The second is the detailed model, which is used to understand the relative impacts of the main distributions and the side lobes on overall accuracy.

\subsubsection{Main Linearised Model}

The inputs to the main model are GMMs fitted to the empirical distributions of the RSRP signals, one mixture for the normal case and one for the faulty case. In our case we were able to achieve good results by using GMMs with four components. The model then processes the input distributions using the linear mathematics as derived in Appendix B. We then compare the distributions at each stage of processing with those from the RNN, in order to validate the mathematical model. We then use the model to understand in depth how the RNN operates on the input distributions to reshape them at each stage of processing. 

\begin{figure*}[!htb] 
  \centering
  \subfloat 
  {\includegraphics[width=5.5cm]{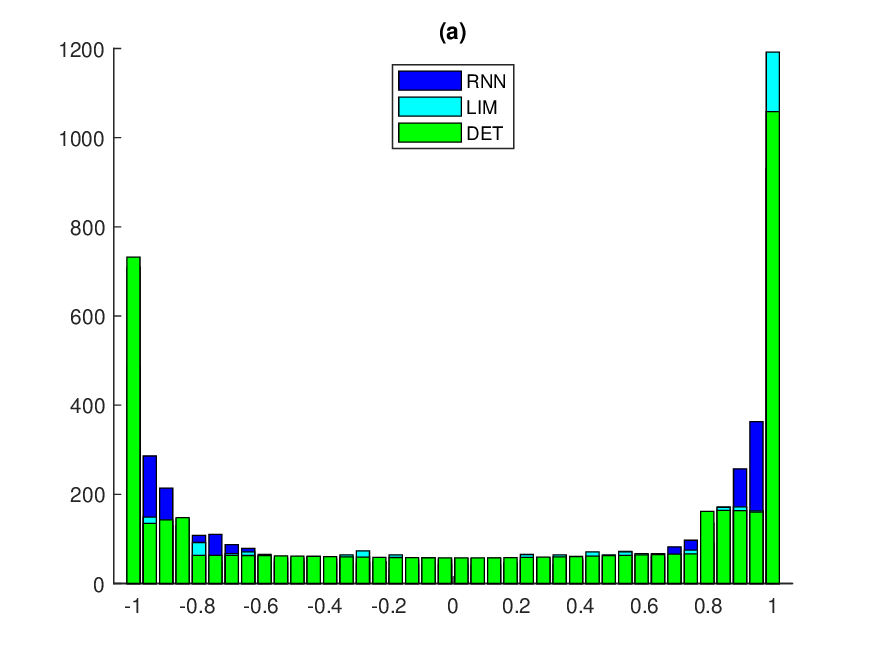}
  \label{fig:1lyr1storder_D4}}
  \subfloat 
  {\includegraphics[width=5.5cm]{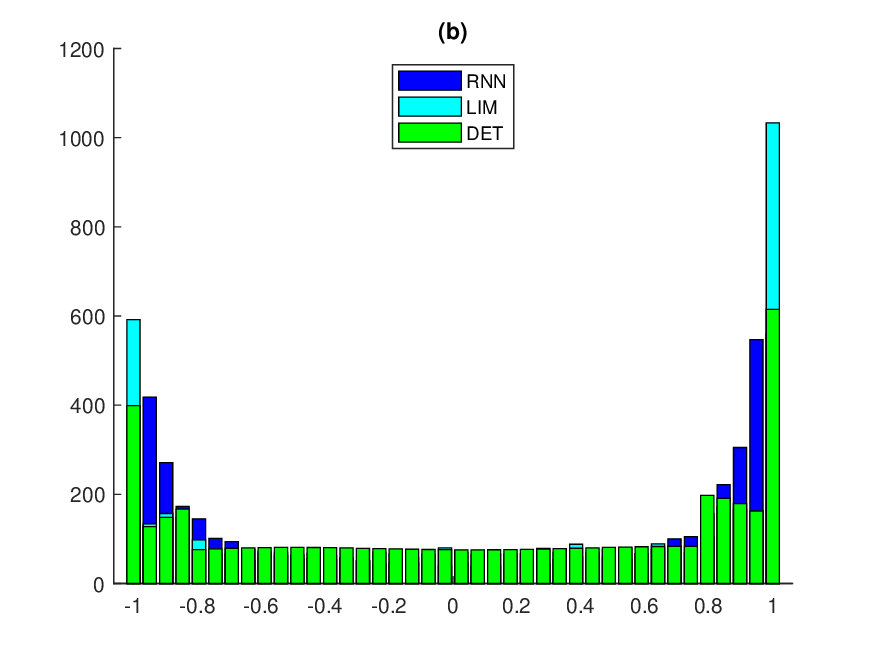}
  \label{fig:2lyr1storder_D42}}
  \subfloat 
  {\includegraphics[width=5.5cm]{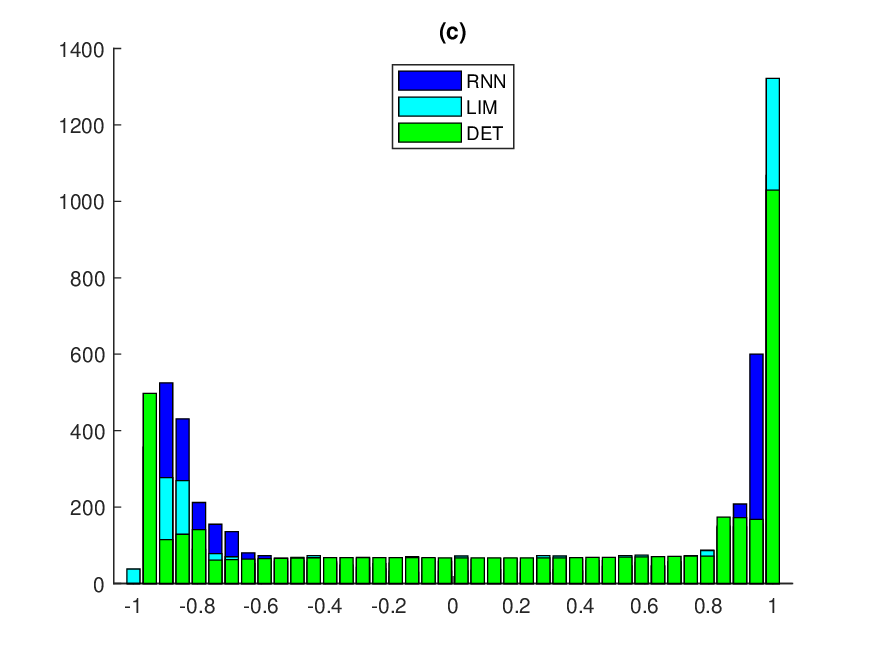}
  \label{fig:3lyr1storder_D43}}
  \newline
  \subfloat 
  {\includegraphics[width=5.5cm]{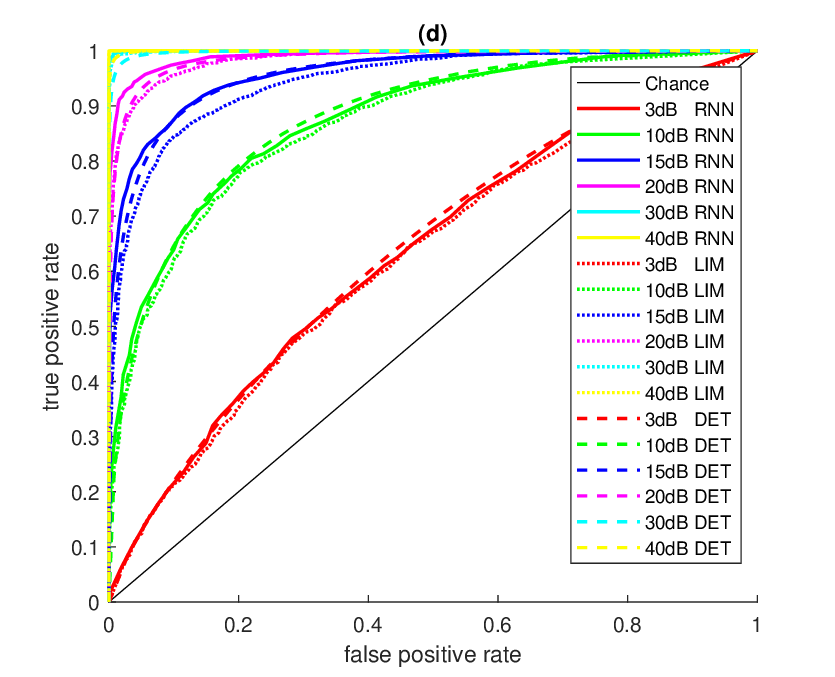}
  \label{fig:1lyr1storder_RoC}}
  \subfloat 
  {\includegraphics[width=5.5cm]{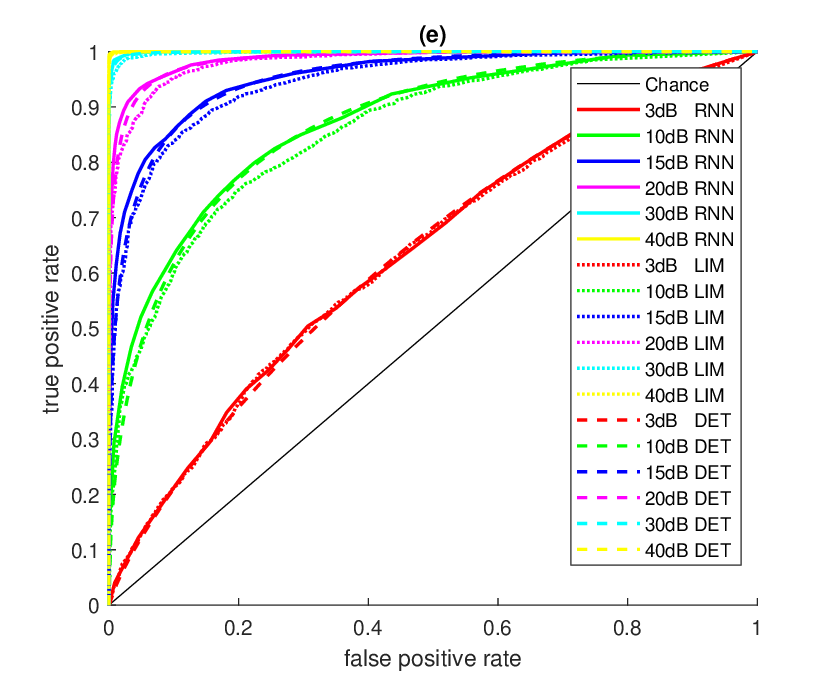}
  \label{fig:2lyr1storder_RoC}}
  \subfloat 
  {\includegraphics[width=5.5cm]{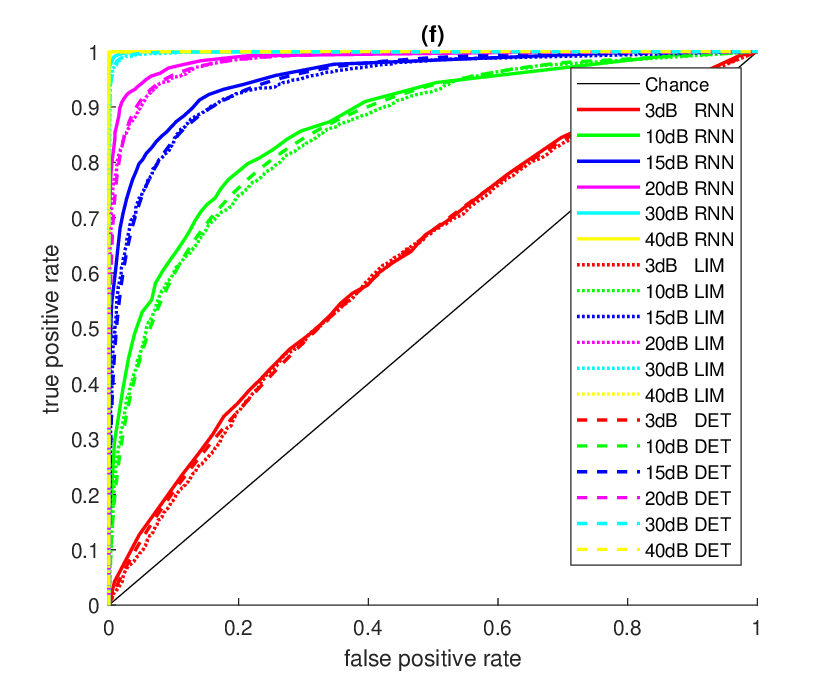}
  \label{fig:3lyr1storder_RoC}}
  \caption{RNN Output Distribution for (a) 1st order one layer, (b) two layers, (c) three layers, and the RoCs for (d) 1st order one layer, (e) two layers, (f) three layers.}
  \label{fig:1storder123layers_D4x_RoCs_RNN}
\end{figure*}

\begin{figure*}[!htb] 
  \centering
  \subfloat
  {\includegraphics[width=5.5cm]{30ir20s2t_antff_8a_1L1_fig210.eps}
  \label{fig:1lyr1storder_D4a}}
  \subfloat 
  {\includegraphics[width=5.5cm]{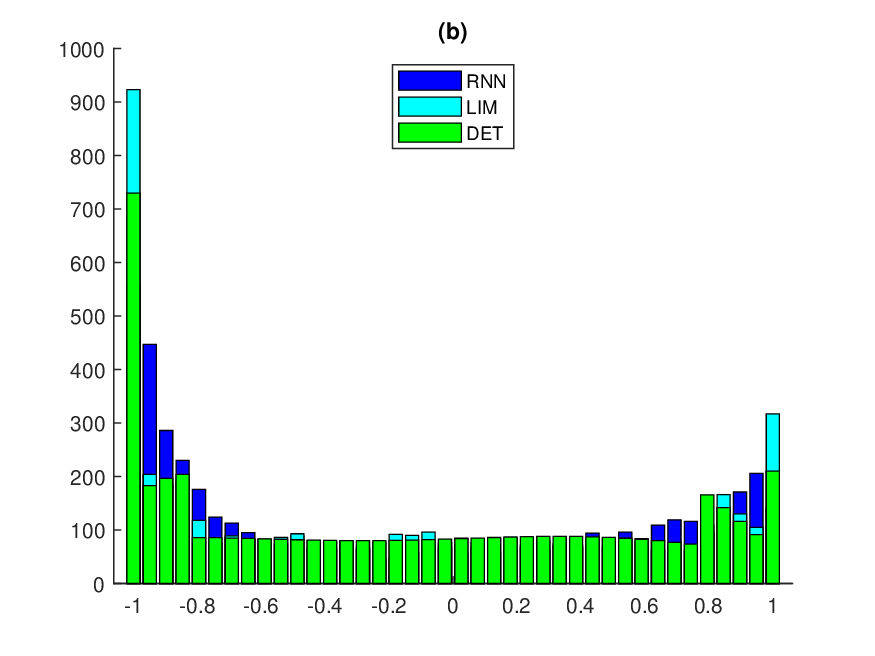}
  \label{fig:1lyr2ndorder_D42}}
  \subfloat 
  {\includegraphics[width=5.5cm]{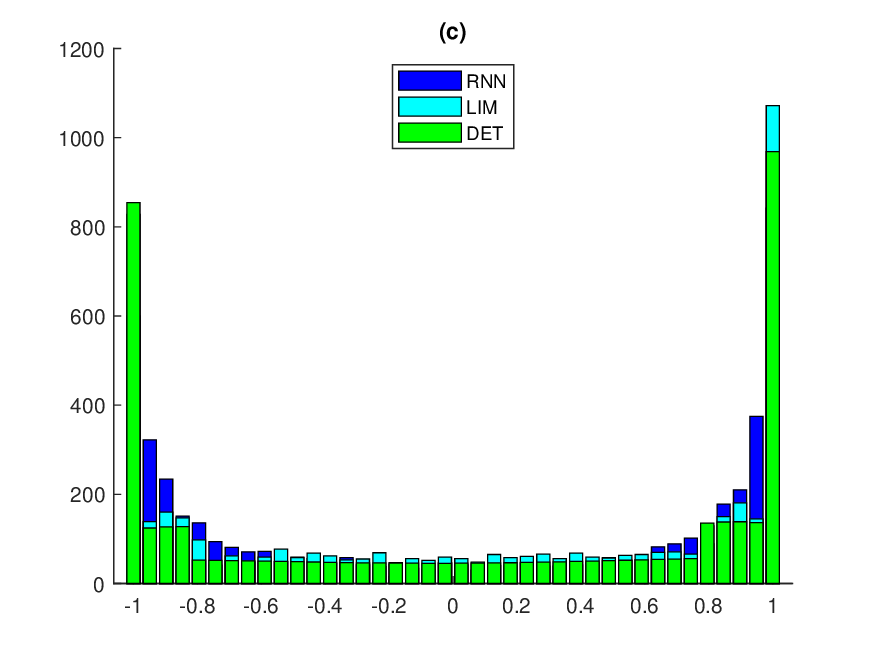}
  \label{fig:1lyr4thorder_D43}}
  \newline
  \subfloat 
  {\includegraphics[width=5.5cm]{30ir20s2t_antff_8a_1L1_fig31.eps}
  \label{fig:1lyr1storder_RoCa}}
  \subfloat 
  {\includegraphics[width=5.5cm]{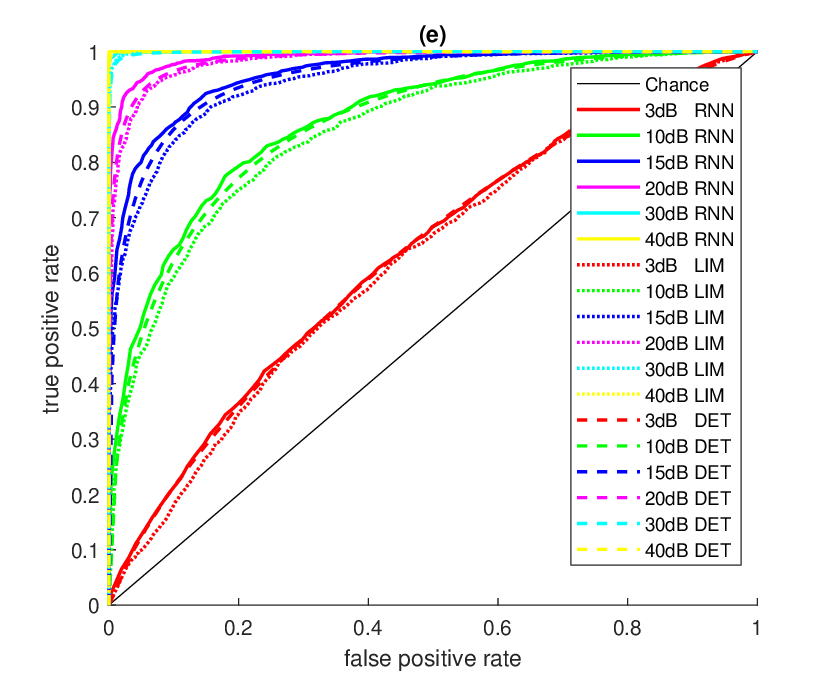}
  \label{fig:1lyr2ndorder_RoC}}
  \subfloat 
  {\includegraphics[width=5.5cm]{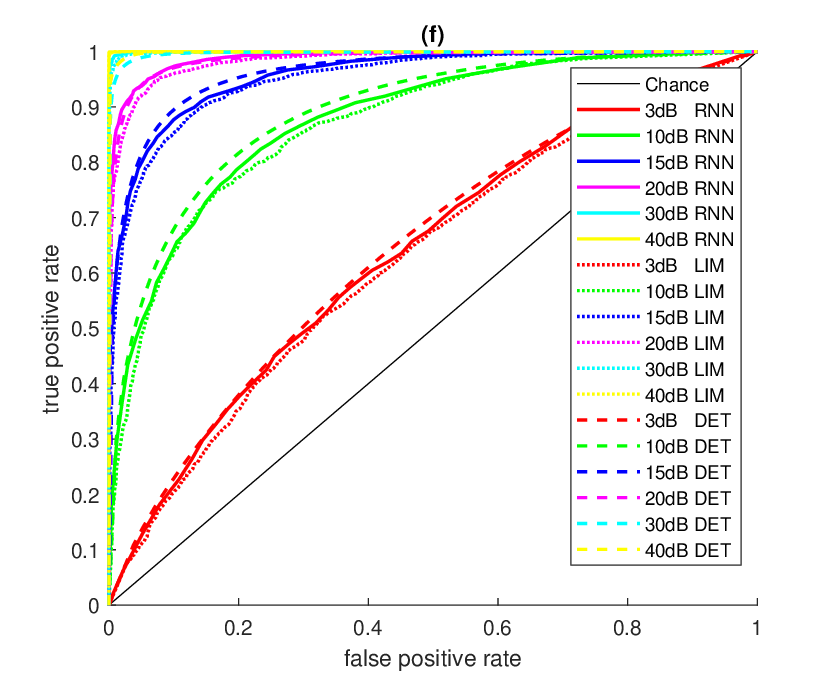}
  \label{fig:1lyr3rdorder_RoC}}
  \caption{Single layer RNN Output Distribution for (a) 1st order, (b) 2nd order, (c) 4th order, and the RoCs for (d) 1st order, (e) 2nd order, (f) 4th order.}
  \label{fig:1layer124orders_D4x_RoCs_RNN}
\end{figure*}

\subsubsection{Detailed Model}

The detailed model takes the mean and variance of the $D_0$ distributions $D_0^{[N]}$ and $D_0^{[F]}$ produced by the main model, and applies the linearised mathematics to generate the $D_k$ distributions. This allows us to view the main and side lobes separately so that we know the mean and standard deviation of each element of the overall distribution and can calculate its contribution to the overall error rate.

For the detailed model we need to take into account two effects. The first is the FSS, namely the combination of $N$ and $F$ statuses in the input sequence as explained above. The second is the LSS, which comprises a set of pairs of $g(\cdot)$ and $r(\cdot)$ values.

These two effects are handled by producing a separate distribution for each combination of FSS and LSS. Each individual distribution is weighted by the product of the relative frequencies of occurrence of the specific FSS and LSS respectively, based on sample counts taken from the main model. The results are then summed to give a combined distribution per FSS.

\section{Validation and Discussion of Results} \label{sec:validation}

\begin{figure*}[!htb] 
  \centering
  \subfloat 
  {\includegraphics[width=5.5cm]{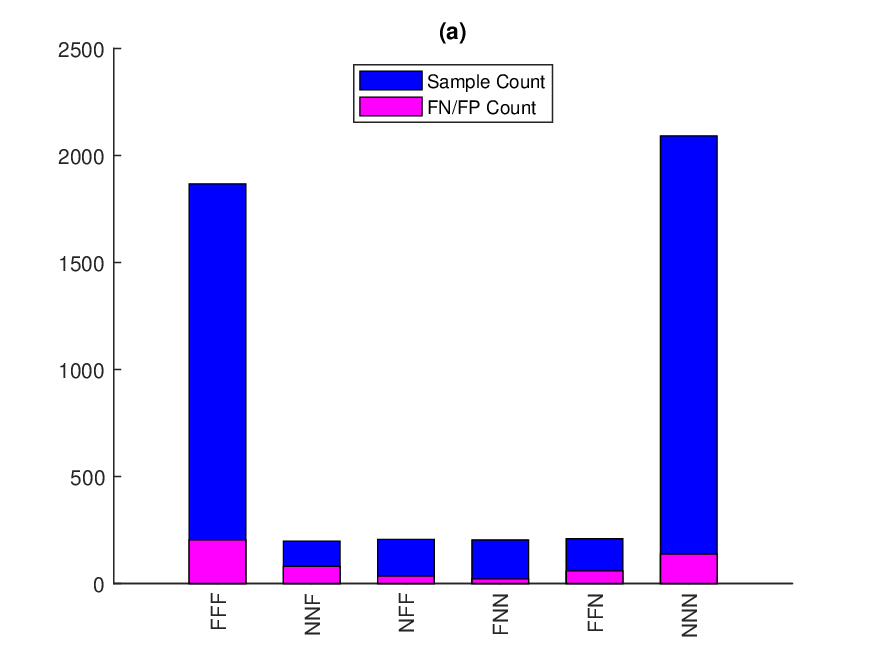}
  \label{fig:1lyr1storder_MD_PSL}}
  \subfloat 
  {\includegraphics[width=5.5cm]{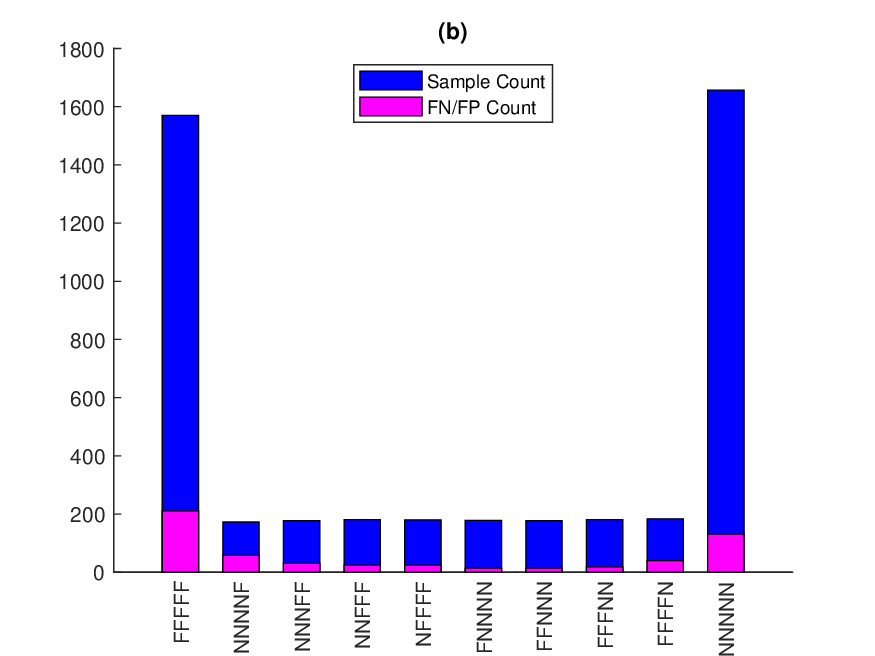}
  \label{fig:2lyr1storder_MD_PSL}}
  \subfloat
  {\includegraphics[width=5.5cm]{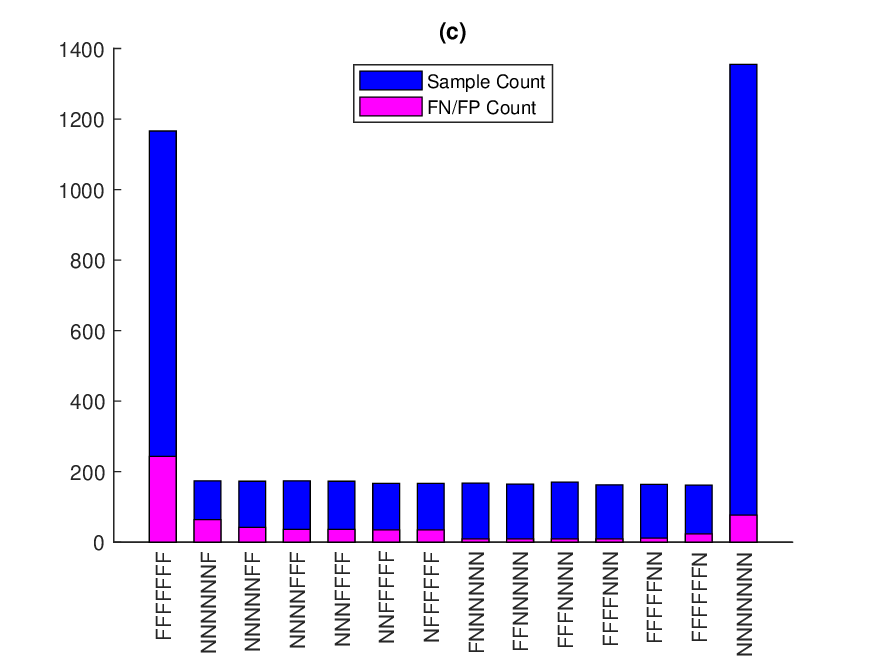}
  \label{fig:3lyr1storder_MD_PSL}}
  \caption{Main Distribution and Principal Sidelobes for (a) single layer, (b) 2 layers, (c) 3 layers.}
  \label{fig:1stOrder123Layers_MNPSL_RNN}
\end{figure*}

\begin{figure*}[!htb] 
  \centering
  \subfloat 
  {\includegraphics[width=5.5cm]{30ir20s2t_antff_8a_1L1_fig307.eps}
 \label{fig:1lyr1storder_MD_PSLa}}
  \subfloat 
  {\includegraphics[width=5.5cm]{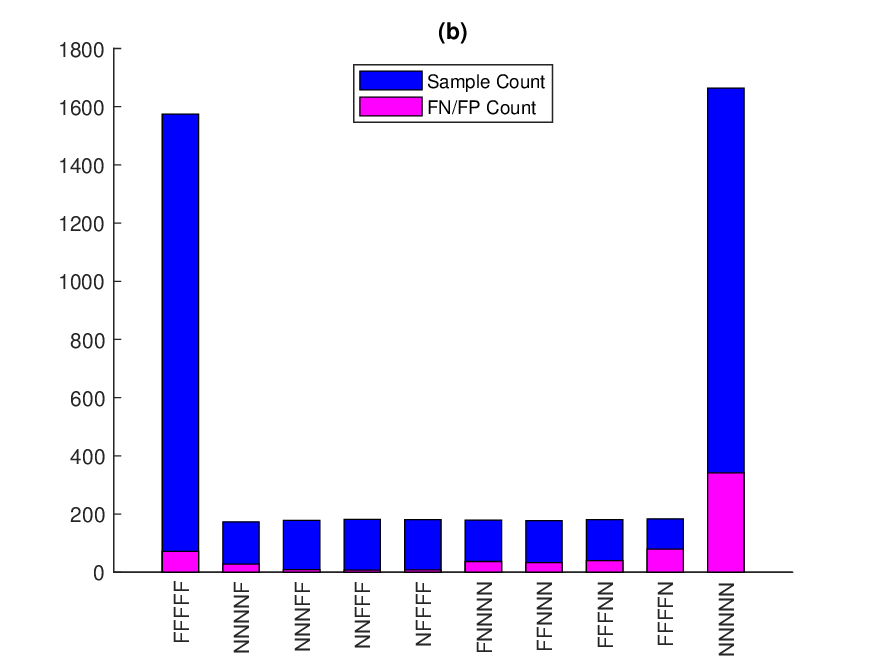}
  \label{fig:1lyr2ndorder_MD_PSL}}
  \subfloat
  {\includegraphics[width=5.5cm]{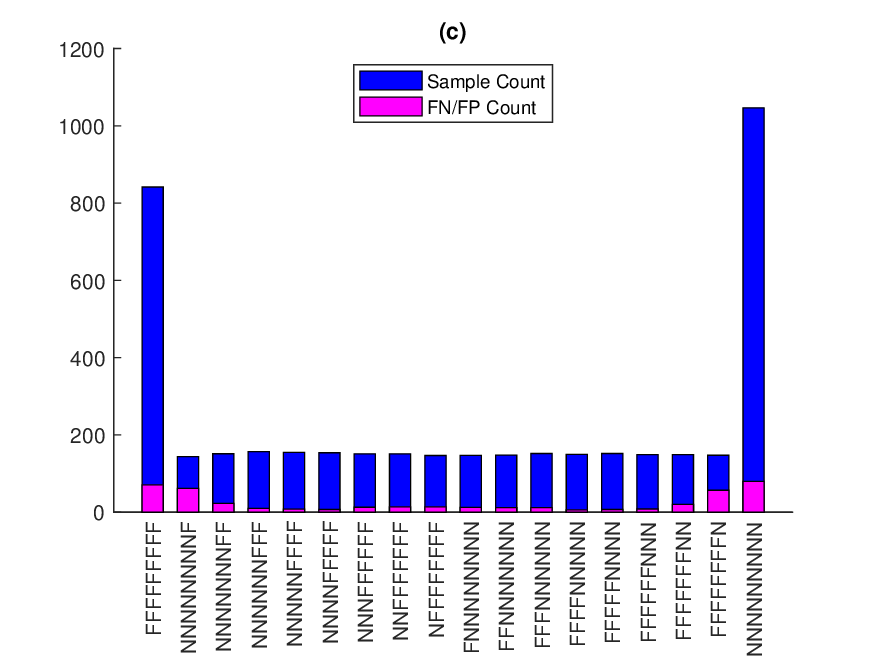}
  \label{fig:1lyr4thorder_MD_PSL}}
  \caption{Main Distribution and Principal Sidelobes for single layer (a) 1st order, (b) 2nd order, and (c) 4th order.}
  \label{fig:1lyr124Orders_MNPSL_RNN}
\end{figure*}

To validate the models, we monitored the internal state vectors of the RNN and compared their values against the equivalents predicted by the linear model and the detailed model. The results of the comparison between the final stages of the RNN and the two models (the $D_k$ distributions and the Receiver Operating Curves)  are shown for multilayer RNN configurations in Fig. \ref{fig:1storder123layers_D4x_RoCs_RNN}  and for different RNN orders in Fig. \ref{fig:1layer124orders_D4x_RoCs_RNN}. We see that there is generally a good fit between the outputs from both models and the RNN outputs, with minor differences due to the approximations we have made in developing the model. A slight mismatch at the shoulders of the D4 distributions is due to the approximation of the smooth tanh curve by linear line segments (which could be arbitrarily reduced by using more line segments). Small differences between the model and RNN outputs at each layer, due to this effect and also arising from the approximations made by limiting the numbers of terms in the feedback transformations, are proportionately larger for low values of fault impact.  Hence for smaller fault impacts we see a relatively larger gap on the RoC curves between the models' predictions and the RNN performance.

We then used the detailed model to study the main and sidelobes in more depth, by plotting the error counts (FN/FP) for each of the main and side lobe distributions. Fig \ref{fig:1stOrder123Layers_MNPSL_RNN}(a) gives these results for the single layer configuration. 

This shows that in relation to $D_{0}$\ there has been an additional decrease in the main lobe error counts due to the effect of the RNN in reducing the variances of the main distributions in relation to the separation between their means. This is counteracted, however, by the false predictions arising from the principal sidelobes, in particular $D_{1}^{[NNF]}$ and $D_{1}^{[FFN]}$. The overall result is that the fault detection accuracy shows a small improvement relative to the output of spatial averaging stage.

Figures \ref{fig:1stOrder123Layers_MNPSL_RNN}(b) gives the results for the two layer case, and  \ref{fig:1stOrder123Layers_MNPSL_RNN}(c) provides the same data for three layers. Comparing the two layer results with those for the single layer configuration, it can be seen that the main lobes include fewer samples and so have become smaller, whereas the remaining samples are distributed across a larger number of sidelobes. The main lobe error counts have fallen slightly but the total error count arising from the sidelobes has increased. The histogram for the third layer shows a similar result. This explains the decreasing improvements in accuracy as we introduce further hidden layers - reductions in the error rate contributed by the main lobes are offset by additional errors contributed by a larger number of sidelobes. We saw the same effect with the higher order configurations, as can be seen from Figs \ref{fig:1lyr124Orders_MNPSL_RNN}(a)-\ref{fig:1lyr124Orders_MNPSL_RNN}(c).

Hence this is the explanation for the limited improvements achieved as a result of either adding further hidden layers to the RNN or introducing higher order internal processing: reductions in the main lobe error counts are counteracted by the production of additional sidelobes each contributing errors.

\section{Further Work}\label{sec:furtherwork}
Potential areas for further work could include study of the applicability of the linearising methods described in this paper to  similar types of NN such as Long Term Short Term Networks (LSTMs) and Gated Recurrent Units (GRUs), as well as looking at the feasibility of extending the method to provide insights into deeper NNs. Extension of the work to other application domains could take place by manually fitting the components of the GMM to the input data distribution, as we have done, or potentially by using the Expectation Maximisation (EM) algorithm. The latter approach is described by Hastie \emph{et al.} \cite{HastieTibshiraniFriedman2011} 6.8 and 8.5 where the authors describe how the EM algorithm can be used to fit a GMM to an arbitrary input distribution.

\section{Conclusion} \label{sec:conclusion}
We have shown that if we take an RNN configuration and transform its state equations into the pdf domain we can obtain a detailed understanding of how the RNN operates. To do this, we assume only that the inputs can be adequately approximated by a Gaussian mixture, which as we have shown is not anticipated to be a significant limitation. We have used this to develop a new approach, classified as a hybrid model under the DARPA explainable AI (XAI)  categorisation scheme, in which a model running in parallel with the RNN is used to illuminate the interior workings of the RNN. By reshaping the pdf at each stage of processing in the same way as the RNN, it closely tracks the RNN outputs and enables us to explain the reasons for the performance limits of a deep RNN. 

Specifically, we have shown why adding more hidden layers to a high performing RNN with a single hidden layer results in diminishing improvements in accuracy. We demonstrated that this is due to the emergence of unwanted sidelobes in the pdf of the output of each internal layer of the RNN. By using our model to study this effect in detail, we are able to show that each sidelobe introduces  additional classification errors. These then detract from an improvement in the overall error rate which would otherwise have been expected from the addition of further layers. We also demonstrated that the same effect occurs when we increase the order of the RNN, in other words we extend the internal state feedback from one to two or even four prior states.

This means that the operation of a deep RNN operating on any inputs which can be represented by a Gaussian mixture is now fully transparent when viewed in the pdf domain, with potential applicability to other types of NN which store their internal state in order to process sequential data. This leads to useful design insights; we conclude, for example, that in cases of  tightly constrained embedded run time computing resources, such as a devolved RAN application, it could be appropriate to deploy a very simple RNN configuration, with perhaps a single hidden layer, rather than increasing the depth and complexity of the system in search of improved performance.

\section*{Acknowledgement}
We would like to acknowledge the support of the University of Surrey 5GIC/6GIC (http://www.surrey.ac.uk/5gic) members for this work.

\begin {appendices} 

\section{Linearisation of Non-linear Function}

\begin{figure}[!tb]
\centering
\includegraphics[width=8cm]{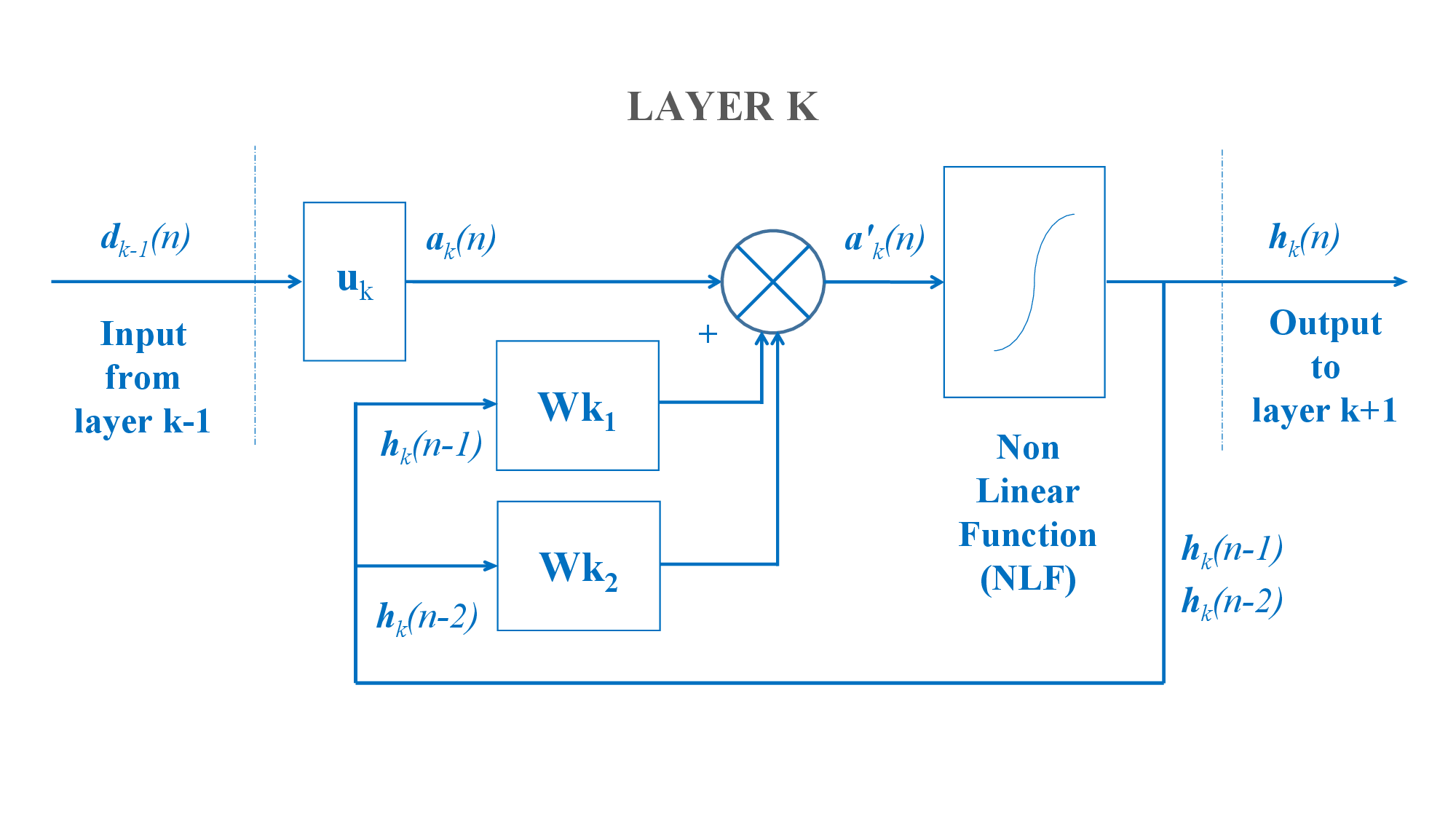}
% where an .eps filename suffix will be assumed under latex, 
% and a .pdf suffix will be assumed for pdflatex; or what has been declared
% via \DeclareGraphicsExtensions.
\caption{Second Order RNN Configuration}
\label{Fig_xAppA}
\end{figure}

A typical NN such as a feedforward NN or a CNN will include one or more non linear functions (NLFs). In the case of an RNN, an additional complication arises because the NLF is inside the loop which feeds back the internal state. We provide an analysis of the RNN case which, in simplified form, can also be applied to the FFNN and CNN cases. 

We consider an RNN with multiple channels, that is, the hidden layer width is greater than 1, so that the internal state at sample $n$ is given by the vector $\textbf{h}(n)$ and the input to the NLF is another vector $\textbf{a}{^\prime}(n)$. When the NLF is applied to a vector, the result is another vector, with the NLF applied elementwise to the input vector to give the output vector. 

We consider the second order case as shown in Fig. \ref{Fig_xAppA}. We can simplify the analysis to cover the first order case or extend it to higher orders using the same logic. We note that for this part of the analysis it is not necessary to assume that the inputs are sampled from any specific probability distribution. In our case the NLF is a tanh$(\cdot)$\ function, but our approach can be applied to other commonly occurring NN NLFs, such as the logistic sigmoid function, where the output increases monotonically as a function of the input.

We first construct a piecewise linear approximation to represent the NLF. First, we define a sequence of pairs of coordinates $\langle (x_1,y_1), (x_2,y_2), \ldots \rangle$ where $x_i<x_{i+1}$ and $y_i<y_{i+1}$. We then utilise the set of all those straight lines which link two points defined by adjacent coordinate pairs, to form an approximation to the non linear element. Let $g_i$ and $r_i$ be the gradient and intercept respectively of the line between the adjacent points $(x_i, y_i)$ and $(x_{i+1}, y_{i+1})$. The applicable line segment may vary for each element of $\textbf{a}{^\prime}(n)$ and may also be different for each sample, so we denote the set of line segments active at sampling instant $n$ by the vectors $\textbf{g}(n)$ and $\textbf{r}(n)$. The output of the NLF at sample $n$ is given by the Hadamard product of the current gain vector and the NLF inputs plus the offset vector, that is:
\begin{equation} \label{eq:tanhlin_0a5}
\textbf{h}(n)=\textbf{g}(n)\odot\textbf{a}{^\prime}(n)
+\textbf{r}(n)
\end{equation} 

\section{Linearisation of Feedback}

In the case of the RNN, however, the situation is complicated by the fact that the NLF is applied to the weighted sum of both the current input $\textbf{a}(n)$ and also one or more previous values of the output of this stage $\textbf{h}(n)$. We, on the other hand, are seeking an approximation to this which operates solely on the inputs, and depends only on a fixed number of previous input values. Consider the second order RNN configuration shown in Fig. \ref{Fig_xAppA}. For a given layer $k$ the input distribution $\textbf{a}{_k}(n)$ is formed by taking the vector $\textbf{d}{_{k-1}}(n)$ sampled from the output distribution of the previous layer $\textbf{D}{_{k-1}}$ and multiplying it by the stage gain $\textbf{u}{_k}$: 

\begin{equation} \label{eq:nlflin_01}
   \textbf{a}{_{k}}(n) = \textbf{u}{_{k}}\textbf{d}{_{k-1}}(n).
\end{equation}

Let the matrix of weights applied to $\textbf{h}(n-j)$ be $ \textbf{W}{_{k,j}}$. For clarity we now drop the subscript $k$ on the understanding that this analysis applies to the general layer $k$. Hence $\textbf{h}(n)$ is given by
\begin{equation} \label{eq:nlflin_02}
\begin{split}
   \textbf{h}(n)\ &= \textbf{g}(n)\odot(\textbf{a}{_\textbf{}}(n) +\textbf{W}{_\textbf{1}}\textbf{h}(n-1)\\
   &+\textbf{W}{_\textbf{2}}\textbf{h}(n-2)) + \textbf{r}(n).\ 
\end{split}
\end{equation}
and therefore 
\begin{equation} \label{eq:nlflin_03}
\begin{split}
    \textbf{h}(n-1)\ &= \textbf{g}(n-1)\odot(\textbf{a}{_\textbf{}}(n-1) +\textbf{W}{_\textbf{1}}\textbf{h}(n-2)\\
    &+\textbf{W}{_\textbf{2}}\textbf{h}(n-3)) + \textbf{r}(n-1)\
\end{split}
\end{equation}
In general
%\begin{scriptsize}
\begin{equation} \label{eq:nlflin_04}
\begin{split}
   \textbf{h}(n-j)\ &= \textbf{g}(n-j)\odot(\textbf{a}{_\textbf{}}(n-j)   +\textbf{W}{_\textbf{1}}\textbf{h}(n-j-1)\\
   &+\textbf{W}{_\textbf{2}}\textbf{h}(n-j-2)) + \textbf{r}(n-j).\
\end{split}
\end{equation}
%\end{scriptsize}
Using the result for $\textbf{h}(n-j)$ in \eqref{eq:nlflin_04}, we can rewrite (\ref {eq:nlflin_02}) as
%\begin{scriptsize}
\begin{equation} \label{eq:nlflin_05}
\begin{split}
   \textbf{h}(n)\ &=\textbf{g}(n)\odot\textbf{a}{_\textbf{}}(n)\\ 
   &+\textbf{r}(n)\\ 
   &+\textbf{g}(n)\odot\textbf{W}{_\textbf{1}}\textbf{g}(n-1)\odot\{\textbf{a}{_\textbf{}}(n-1)\\
   &+\textbf{W}{_\textbf{1}}\textbf{h}(n-2)+\textbf{W}{_\textbf{2}}\textbf{h}(n-3)\} +\textbf{r}(n-1)]\\
   &+\textbf{g}(n)\odot\textbf{W}{_\textbf{2}}[\textbf{g}(n-2)\odot\{\textbf{a}{_\textbf{}}(n-2)\\
   &+\textbf{W}{_\textbf{1}}\textbf{h}(n-3)\\
   &+\textbf{W}{_\textbf{2}}\textbf{h}(n-4)\} +\textbf{r}(n-2)].
    \end{split}  
\end{equation}
%\end{scriptsize}

By continuing to substitute for $\textbf{h}(n-2)$ to $\textbf{h}(n-4)$ in (\ref {eq:nlflin_05}), we arrive at
%\begin{scriptsize}
\begin{equation} \label{eq:nlflin_06}
\begin{split}
   \textbf{h}(n)\ &=\textbf{g}(n)\odot\textbf{a}{_\textbf{}}(n)\\ 
   &+\textbf{r}(n)\\ 
   &+\textbf{g}(n)\odot\textbf{W}{_\textbf{1}}[\textbf{g}(n-1)\odot\{\textbf{a}{_\textbf{}}(n-1)\\
   &+\textbf{W}{_\textbf{1}}\langle \textbf{g}(n-2)\odot(\textbf{a}{_\textbf{}}(n-2)    +\textbf{W}{_\textbf{1}}\textbf{h}(n-3)\\
   &+\textbf{W}{_\textbf{2}}\textbf{h}(n-4)) 
   +\textbf{r}(n-2)\rangle\\
   &+\textbf{W}{_\textbf{2}}\langle \textbf{g}(n-3)\odot(\textbf{a}{_\textbf{}}(n-3)
   +\textbf{W}{_\textbf{1}}\textbf{h}(n-4)\\
   &+\textbf{W}{_\textbf{2}}\textbf{h}(n-5))
   +\textbf{r}(n-3)\rangle\}\\
   &+\textbf{r}(n-1)]\\
   &+\textbf{g}(n)\odot\textbf{W}{_\textbf{2}}[\textbf{g}(n-2)\odot\{\textbf{a}{_\textbf{}}(n-2)\\
   &+\textbf{W}{_\textbf{1}}\langle \textbf{g}(n-3)\odot(\textbf{a}{_\textbf{}}(n-3)
   +\textbf{W}{_\textbf{1}}\textbf{h}(n-4)\\
   &+\textbf{W}{_\textbf{2}}\textbf{h}(n-5))
   +\textbf{r}(n-3)\rangle\\
   &+\textbf{W}{_\textbf{2}}\langle \textbf{g}(n-4)\odot(\textbf{a}{_\textbf{}}(n-4)
   +\textbf{W}{_\textbf{1}}\textbf{h}(n-5)\\
   &+\textbf{W}{_\textbf{2}}\textbf{h}(n-6))\\
   &+\textbf{r}(n-4)\rangle\}\\
   &+\textbf{r}(n-2)].
\end{split}  
\end{equation}
%\end{scriptsize}

We can then continue substituting recursively for $\textbf{h}(n-k)$ as far as we wish. In this case, however, given that the gradient of tanh$(x)$ is less than or equal to 1, and accepting the restriction that $w{_k}< 1$ we can approximate this by ignoring terms beyond second order products of both $g$ and $w$. From the above equation it can be seen that this is equivalent to dropping the terms in $\textbf{h}(n-k)$ such that $k\geq\ 3$. If we eliminate these we have
%\begin{scriptsize}
\begin{equation} \label{eq:nlflin_07}
\begin{split}
   \textbf{h}(n)\ &=\textbf{g}(n)\odot\textbf{a}{_\textbf{}}(n) +\textbf{r}(n)\\
   &+\textbf{g}(n)\odot\textbf{W}{_\textbf{1}}[\textbf{g}(n-1)\odot\{\textbf{a}{_\textbf{}}(n-1)\\
   &+\textbf{W}{_\textbf{1}}\langle \textbf{g}(n-2)\odot\textbf{a}{_\textbf{}}(n-2)    +     \textbf{r}(n-2)\rangle\\
   &+\textbf{W}{_\textbf{2}}\langle \textbf{g}(n-3)\odot\textbf{a}{_\textbf{}}(n-3)
   +\textbf{r}(n-3)\rangle\}\\
   &+\textbf{r}(n-1)]\\
   &+\textbf{g}(n)\odot\textbf{W}{_\textbf{2}}[\textbf{g}(n-2)\odot\{\textbf{a}{_\textbf{}}(n-2)\\
   &+\textbf{W}{_\textbf{1}}\langle \textbf{g}(n-3)\odot\textbf{a}{_\textbf{}}(n-3)
   +\textbf{r}(n-3)\rangle\\
   &+\textbf{W}{_\textbf{2}}\langle \textbf{g(}n-4)\odot\textbf{a}{_\textbf{}}(n-4)
   +    \textbf{r}(n-4)\rangle\}\\
   &+\textbf{r}(n-2)].
\end{split}  
\end{equation}
%\end{scriptsize}

If we expand this out and rearrange we get
\begin{scriptsize}
\begin{equation} \label{eq:nlflin_08}
\begin{split}
   \textbf{h}(n)\ &=\textbf{g}(n)\odot\textbf{n}{_\textbf{}}(n) +\textbf{r}(n)\\
   &+\textbf{g}(n)\odot\textbf{W}{_\textbf{1}}\textbf{g}(n-1)\odot\textbf{a}{_\textbf{}}(n-1)\\
   &+\textbf{g}(n)\odot\textbf{W}{_\textbf{1}}\textbf{W}{_\textbf{1}}\textbf{g}(n-1)\odot\langle    \textbf{g}(n-2)\odot\textbf{a}{_\textbf{}}(n-2) + \textbf{r}(n-2)\rangle\\
   &+\textbf{g}(n)\odot\textbf{W}{_\textbf{1}}\textbf{W}{_\textbf{2}}\textbf{g}(n-1)\odot\langle    \textbf{g}(n-3)\odot\textbf{a}{_\textbf{}}(n-3) + \textbf{r}(n-3)\rangle\\
   &+\textbf{g}(n)\odot\textbf{W}{_\textbf{1}}\textbf{r}(n-1)\\    
   &+\textbf{g}(n)\odot\textbf{W}{_\textbf{2}}\textbf{g}(n-2)\odot\textbf{a}{_\textbf{}}(n-2)\\
   &+\textbf{g}(n)\odot\textbf{W}{_\textbf{2}}\textbf{W}{_\textbf{1}}\textbf{g}(n-2)\odot\langle
   \textbf{g}(n-3)\odot\textbf{a}{_\textbf{}}(n-3)+ \textbf{r}(n-3)\rangle\\
   &+\textbf{g}(n)\odot\textbf{W}{_\textbf{2}}\textbf{W}{_\textbf{2}}.\textbf{g}(n-2)\odot\langle    \textbf{g}(n-4)\odot\textbf{a}{_\textbf{}}(n-4) + \textbf{r}(n-4)\rangle\\
   &+\textbf{g}(n)\odot\textbf{W}{_\textbf{2}}\textbf{r}(n-2)
\end{split}  
\end{equation}
\end{scriptsize}

The values of $\textbf{g}(\cdot)$ and $\textbf{r}(\cdot)$ are different for each line segment, and the selection of active line segment is determined by each element of $\textbf{a}{^\prime}(n)$. This gives a set of combinations of pairs of $g(\cdot)$ and $r(\cdot)$ values for each channel (assuming all the $\textbf{W}$ matrices are diagonal) which we refer to as the line segment sequence. In the second order case, where we consider the current input and the previous four inputs, this sequence will consist of five pairs of $g(\cdot)$ and $r(\cdot)$ values:
\begin{equation*}
\langle (g_0^{[l]},r_0^{[l]}), (g_1^{[l]},r_1^{[l]})... (g_4^{[l]},r_4^{[l]}) \rangle
\end{equation*} where $[l]$ identifies the line segment sequence for a given channel, and such that the pair ($g_j^{[l]} , r_j^{[l]})$ maps onto the pair $(g(n-j), r(n-j)$).

For multiple channels, in general the line segment sequence will differ for each channel so we identify the set of line segment sequences by $\textbf{[L]}$.

If we then define coefficient vectors $\boldsymbol{\alpha}{_\textbf{0}}{^{\textbf{[L]}}}$ - $\boldsymbol{\alpha}{_\textbf{4}}{^{\textbf{[L]}}}$ and an offset vector $\boldsymbol{\beta}{^{\textbf{[L]}}}$,  we can rewrite (\ref {eq:nlflin_08}) as
\begin{equation} \label{eq:nlflin_09}
\begin{split}
    \textbf{h}(n) &= \boldsymbol{\alpha}{_\textbf{0}}{^{\textbf{[L]}}}\textbf{a}(n) 
   +\boldsymbol{\alpha}{_\textbf{1}}{^{\textbf{[L]}}}\textbf{a}(n-1)\\
   &+\boldsymbol{\alpha}{_\textbf{2}}{^{\textbf{[L]}}}\textbf{a}(n-2) 
   +\boldsymbol{\alpha}{_\textbf{3}}{^{\textbf{[L]}}}\textbf{a}(n-3)
   +\boldsymbol{\beta}{^{\textbf{[L]}}}   
\end{split}
\end{equation} 
where 
\begin{equation} \label{eq:nlflin_10}
\begin{split}
    \boldsymbol{\alpha}{_\textbf{0}}{^{\textbf{[L]}}} &=\textbf{g}{_\textbf{0}}\\
    \boldsymbol{\alpha}{_\textbf{1}}{^{\textbf{[L]}}}&=\textbf{g}{_\textbf{0}}\odot\textbf{W}{_\textbf{1}}\textbf{g}{_\textbf{1}}\\
    \boldsymbol{\alpha}{_\textbf{2}}{^{\textbf{[L]}}}&=\textbf{g}{_\textbf{0}}\odot\textbf{g}{_\textbf{1}}\odot\textbf{W}{_\textbf{1}}\textbf{W}{_\textbf{1}} \textbf{g}{_\textbf{2}}+\textbf{g}{_\textbf{0}}\odot\textbf{W}{_\textbf{2}}\textbf{g}{_\textbf{2}}\\
    \boldsymbol{\alpha}{_\textbf{3}}{^{\textbf{[L]}}}&=\textbf{g}{_\textbf{0}}\odot\textbf{g}{_\textbf{1}}\odot\textbf{W}{_\textbf{1}}\textbf{W}{_\textbf{2}}  \textbf{g}{_\textbf{3}}+\textbf{g}{_\textbf{0}}\odot\textbf{g}{_\textbf{2}}\odot\textbf{W}{_\textbf{2}}\textbf{W}{_\textbf{1}}\textbf{g}{_\textbf{3}}\\
    \boldsymbol{\alpha}{_\textbf{4}}{^{\textbf{[L]}}}&=\textbf{g}{_\textbf{0}}\odot\textbf{g}{_\textbf{2}}\odot\textbf{W}{_\textbf{2}}\textbf{W}{_\textbf{2}} \textbf{g}{_\textbf{4}}
\end{split}
\end{equation}
and
\begin{equation} \label{eq:nlflin_11}
\begin{split}
    \boldsymbol{\beta}{^{\textbf{[L]}}}&=\textbf{r}{_\textbf{0}}+\textbf{g}{_\textbf{0}}\odot\textbf{W}{_\textbf{1}}\textbf{r}{_\textbf{1}}\\ &+[\textbf{g}{_\textbf{0}}\odot\textbf{W}{_\textbf{1}}\textbf{W}{_\textbf{1}}\textbf{g}{_\textbf{1}}
+\textbf{g}{_\textbf{0}}\odot\textbf{W}{_\textbf{2}}]\textbf{r}{_\textbf{2}}\\
&+[\textbf{g}{_\textbf{0}}\odot\textbf{W}{_\textbf{1}}\textbf{W}{_\textbf{2}}\textbf{g}{_\textbf{1}}
+\textbf{g}{_\textbf{0}}\odot\textbf{W}{_\textbf{2}}\textbf{W}{_\textbf{1}}\textbf{g}{_\textbf{2}}]\textbf{r}{_\textbf{3}}\\
&+\textbf{g}{_\textbf{0}}\odot\textbf{W}{_\textbf{2}}\textbf{W}{_\textbf{2}}\textbf{g}{_\textbf{2}}]\textbf{r}{_\textbf{4}}.
\end{split}
\end{equation}

For first order systems we can set $\textbf{W}{_\textbf{2}}$ to zero, so these equations reduce to
\begin{equation} \label{eq:nlflin_09c}
\begin{split}
\textbf{h}(n)\
  &= \boldsymbol{\alpha}{_\textbf{0}}{^{\textbf{[L]}}}\textbf{a}(n) 
   +\boldsymbol{\alpha}{_\textbf{1}}{^{\textbf{[L]}}}\textbf{a}(n-1)
\\&+\boldsymbol{\alpha}{_\textbf{2}}{^{\textbf{[L]}}}\textbf{a}(n-2) 
+\boldsymbol{\beta}{^{\textbf{[L]}}}   
\end{split}
\end{equation}
where 
\begin{equation} \label{eq:nlflin_15}
\begin{split}
\boldsymbol{\alpha}{_\textbf{0}}{^{\textbf{[L]}}} &=\textbf{g}{_\textbf{0}}\\
 \boldsymbol{\alpha}{_\textbf{1}}{^{\textbf{[L]}}}&=\textbf{g}{_\textbf{0}}\odot\textbf{W}{_\textbf{1}}\textbf{g}{_\textbf{1}}
\\ \boldsymbol{\alpha}{_\textbf{2}}{^{\textbf{[L]}}}&=\textbf{g}{_\textbf{0}}\odot\textbf{g}{_\textbf{1}}\odot\textbf{W}{_\textbf{1}}\textbf{W}{_\textbf{1}}   \textbf{g}{_\textbf{2}}
\end{split}
\end{equation}
and
\begin{equation} \label{eq:nlflin_16}
   \boldsymbol{\beta}{^{\textbf{[L]}}}=\textbf{r}{_\textbf{0}}+\textbf{g}{_\textbf{0}}\odot\textbf{W}{_\textbf{1}}\textbf{r}{_\textbf{1}} \\ +[\textbf{g}{_\textbf{0}}\odot\textbf{W}{_\textbf{1}}\textbf{W}{_\textbf{1}}\textbf{g}{_\textbf{1}}]\textbf{r}{_\textbf{2}}.
\end{equation}

For fourth order systems we need to consider the current input and the previous eight inputs. We can apply the same approach as above, however, to arrive at similar results, giving nine coefficient vectors $\boldsymbol{\alpha}{_\textbf{0}}{^{\textbf{[L]}}}$ - $\boldsymbol{\alpha}{_\textbf{8}}{^{\textbf{[L]}}}$ plus the offset vector $\boldsymbol{\beta}{^{\textbf{[L]}}}$. 

\end {appendices}

\ifCLASSOPTIONcaptionsoff
  \newpage
\fi

%\bibliographystyle{IEEEtran}
%\bibliography{ref}
%\bibliography{references}

\bibliographystyle{IEEEtran}
%\bibliography{IEEEabrv,../full_list}
\bibliography{IEEEabrv,full_list}

\newpage %prevent biographies from following on bibliography
\begin{IEEEbiography}
%[{\includegraphics[width=1in,height=1.25in,clip,keepaspectratio]{dmulvey}}]
%[{\includegraphics[height=1.25in,keepaspectratio]{dmulvey.eps}}]
[{\includegraphics[height=1.25in,keepaspectratio]{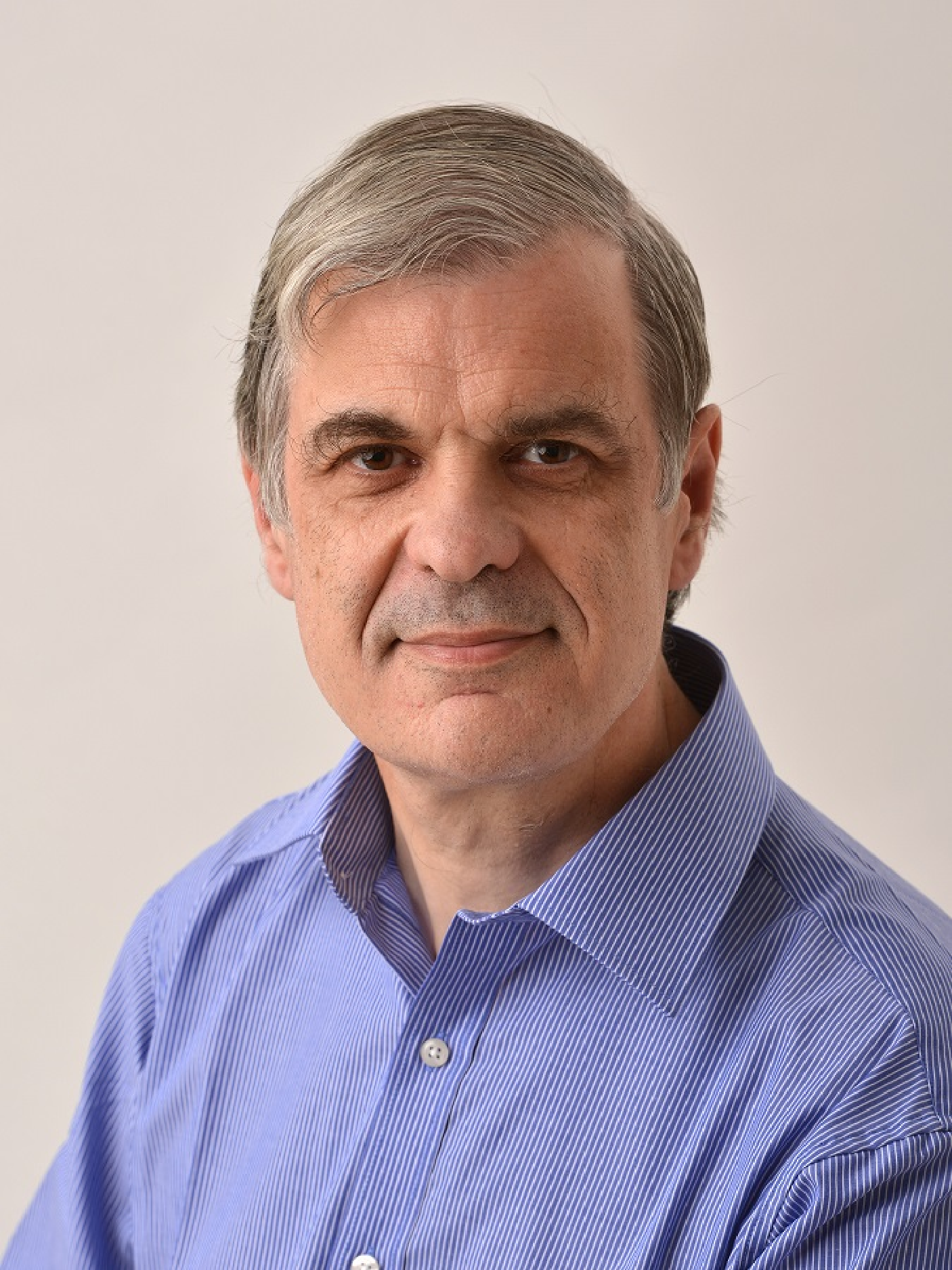}}]
{David Mulvey}received his B.A. degree from the University of Cambridge in 1978, and an M.Sc. degree in Digital Control Systems from Bristol University in 1986. After a career in systems engineering for telecommunications and mission-critical control, he is currently studying towards a Ph.D. degree at the 5G/6G Innovation Centre in the  University of Surrey, UK.
His main research interests include the application of deep learning techniques to the management of cellular radio networks. 
He is a Fellow of the IET.
\end{IEEEbiography}

\vspace{-10 mm} %reduce spacing between biographies

\begin{IEEEbiography}
%[{\includegraphics[width=1in,height=1.25in,clip,keepaspectratio]{ch_foh.eps}}]
[{\includegraphics[height=1.25in,keepaspectratio]{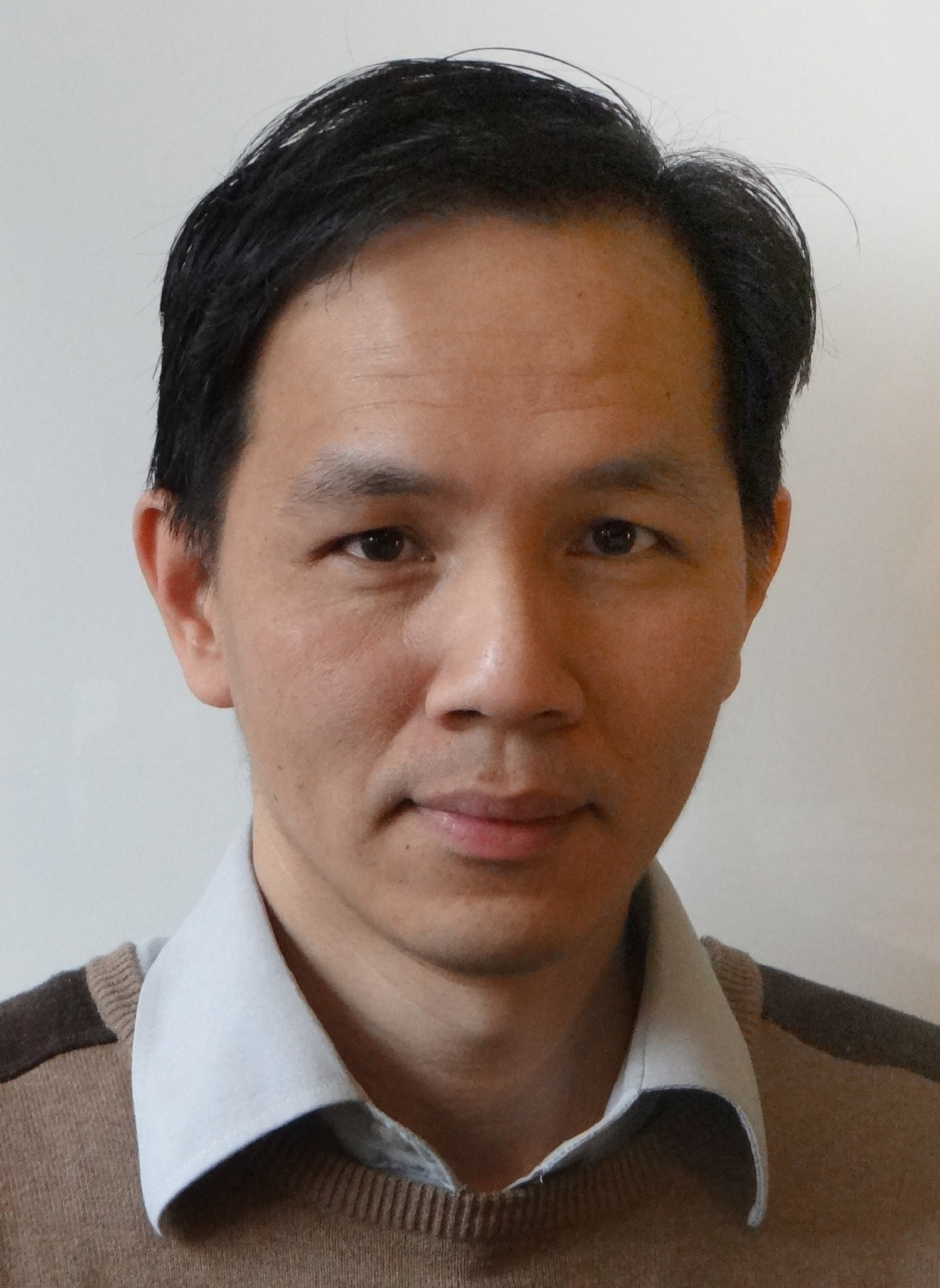}}]
{Chuan Heng Foh}received his M.Sc. degree from Monash University, Australia in 1999 and Ph.D. degree from the University of Melbourne, Australia in 2002. After his PhD, he spent 6 months as a Lecturer at Monash University in Australia. In December 2002, he joined Nanyang Tchnological University, Singapore as an Assistant Professor until 2012. He is now a Senior Lecturer at the University of Surrey. His research interests include protocol design and performance analysis of various computer networks including wireless local area and mesh networks, mobile ad hoc and sensor networks, Internet of Things, 5G networks, and data center networks. He has authored or coauthored over 100 refereed papers in international journals and conferences. In 2015-2017, he served as the Vice-Chair (Europe/Africa) of IEEE Technical Committee on Green Communications and Computing (TCGCC). He is currently an Associate Editor for IEEE Access, IEEE Wireless Communications, and International Journal of Communications Systems. He is a senior member of IEEE.
\end{IEEEbiography}

%\newpage
%
\vspace{-10 mm} %reduce spacing between biographies

\begin{IEEEbiography}
%[{\includegraphics[width=1in,height=1.25in, clip,keepaspectratio]{maimran3}}]
[{\includegraphics[height=1.25in,keepaspectratio]{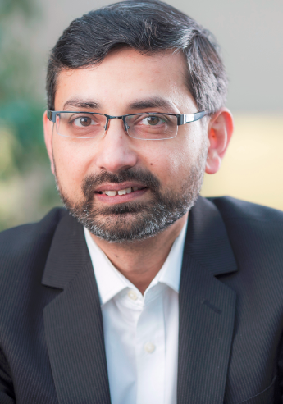}}]
{Muhammad Imran}is the Vice Dean Glasgow College UESTC and Professor of Communication Systems in the School of Engineering at the University of Glasgow.  He was awarded his M.Sc. (Distinction) and Ph.D. degrees from Imperial College London, U.K., in 2002 and 2007, respectively. He is an Affiliate Professor at the University of Oklahoma, USA and a visiting Professor at 5G Innovation Centre, University of Surrey, UK. He has over 18 years of combined academic and industry experience, working primarily in the research areas of cellular communication systems.  He has been awarded 15 patents, has authored/co-authored over 300 journal and conference publications, and has been principal/co-principal investigator on over £6 million in sponsored research grants and contracts. He has supervised 30+ successful PhD graduates.
He has an award of excellence in recognition of his academic achievements, conferred by the President of Pakistan. He was also awarded IEEE Comsoc’s Fred Ellersick award 2014, FEPS Learning and Teaching award 2014, Sentinel of Science Award 2016. He was twice nominated for Tony Jean’s Inspirational Teaching award. He is a shortlisted finalist for The Wharton-QS Stars Awards 2014, QS Stars Reimagine Education Award 2016 for innovative teaching and VC’s learning and teaching award in University of Surrey. He is a senior member of IEEE and a Senior Fellow of Higher Education Academy (SFHEA), UK.

\end{IEEEbiography}

\vspace{-13 mm} %reduce spacing between biographies

\begin{IEEEbiography}
%[{\includegraphics[width=1in,height=1.25in,clip,keepaspectratio]{rtafazolli3}}]
[{\includegraphics[height=1.25in,keepaspectratio]{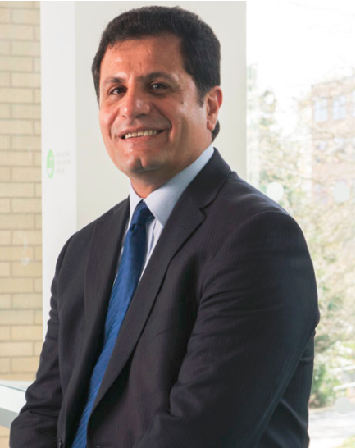}}]
{Rahim Tafazolli}is currently a Professor and the Director of the 5G/6G Innovation Center, Institute for Communications Systems, University of Surrey, U.K. He has authored over 500 research papers in refereed
journals, international conferences, and has been an invited speaker. He edited two books Technologies for Wireless Future (Wiley, 2004 and 2006). He was appointed a Fellow of the Wireless World Research Forum in 2011, in recognition of his personal contribution to the wireless world. He is the head of one of Europe’s leading research groups.
\end{IEEEbiography}

%\begin{IEEEbiography}[{\includegraphics[width=1in,height=1.25in,clip,keepaspectratio]{picture}}]{John Doe}
%\blindtext
%\end{IEEEbiography}
\end{document}